\documentclass[journal]{IEEEtran}
\usepackage{amsmath,amsfonts}
\usepackage{algorithmic}
\usepackage{array}
\usepackage[caption=false,font=normalsize,labelfont=sf,textfont=sf]{subfig}
\usepackage{textcomp}
\usepackage{stfloats}
\usepackage{url}
\usepackage{verbatim}
\usepackage{graphicx}
\def\BibTeX{{\rm B\kern-.05em{\sc i\kern-.025em b}\kern-.08em
    T\kern-.1667em\lower.7ex\hbox{E}\kern-.125emX}}
\usepackage{balance}
\usepackage[backend=biber,style=numeric, sorting=none]{biblatex}
\begin{document}
\title{Fast frequency reconstruction using Deep Learning for event recognition in ring laser data}

\author{Giuseppe Di Somma$^{2}$ Giorgio Carelli$^{1,2}$ Angela D.V. Di Virgilio$^{2}$ Francesco Fuso$^{1,2}$ Enrico Maccioni$^{1,2}$ Paolo Marsili$^{1,2}$ 

$^1$ Dipartimento di Fisica, Universit\`a di Pisa, largo B. Pontecorvo 3, I-56127 Pisa, Italy \\
$^2$ Istituto Nazionale di Fisica Nucleare (INFN), sez. di Pisa, largo B. Pontecorvo 3, I-56127 Pisa, Italy \\

}

\maketitle

\begin{abstract}

The reconstruction of a frequency with minimal delay from a sinusoidal signal is a common task in several fields; for example Ring Laser Gyroscopes, since their output signal is a beat frequency. While conventional methods require several seconds of data, we present a neural network approach capable of reconstructing frequencies of several hundred Hertz within approximately 10 milliseconds. This enables rapid trigger generation. The method outperforms standard Fourier-based techniques, improving frequency estimation precision by a factor of 2 in the operational range of GINGERINO, our Ring Laser Gyroscope.\\
In addition to fast frequency estimation, we introduce an automated classification framework to identify physical disturbances in the signal, such as laser instabilities and seismic events, achieving accuracy rates between 99\% and 100\% on independent test datasets for the seismic class. These results mark a step forward in integrating artificial intelligence into signal analysis for geophysical applications.

\end{abstract}


\section{Introduction}

Sagnac interferometers are commonly used for measuring absolute angular velocities, such as Earth's rotation \cite{ADV2022, schreiber2023}. The Sagnac effect is a physical phenomenon in which two light beams traveling in opposite directions around a closed loop detect a difference in path length due to the rotation of the system \cite{sagnac1913}, and is extensively utilized in inertial navigation \cite{BigBook, King}.\\
Among that Sagnac Interferometers, the large-frame Ring Laser Gyroscope (RLG) is the most sensitive device. Sensitivities on the order of prad/s, coupled with broad dynamic range and continuous operation, have been demonstrated \cite{uno, due, DiSomma}.\\
The equation relating the frequency encoded in the RLG interference signal to the cavity rotation rate is \cite{AA19}:
\begin{equation}
    f_s = \frac{4 \, \Omega \, A}{P \, \lambda} \cos\theta
    \label{eq:SagnacFrequency}
\end{equation}
where $ A $  is the area enclosed by the ring laser cavity, $ P$ is the perimeter of the cavity, $ \lambda $ is the wavelength of the laser light, $\theta$ is the angle between the ring area vector and the $\Omega$ axis of rotation. If the gyroscope is positioned horizontally with respect to the Earth's surface, the angle $\theta $ corresponds to the colatitude.\\
GINGERINO is an RLG with a 3.6 m square cavity, operating at INFN’s underground Gran Sasso laboratory \cite{AA24, AA28}. It has been built in order to validate the Gran Sasso laboratory for the installation of the GINGER experiment \cite{GINGER, StatusGINGER}, based on an array of RLG, currently under construction. Large frame RLGs sensitivity enables the detection of geophysical signals, including polar motion, tides, and seismic events, and allow fundamental physics tests \cite{DiSomma:2024gxf, Capozziello}.\\
This is the first time that Neural Networks (NN) have been applied to RLG data analysis. The signal of interest is the beat note of the two RLG beams, recombined outside the cavity, from which the frequency must be reconstructed. Typically, frequency reconstruction is performed offline using an analytic function based on the Hilbert transform, requiring few seconds of delay \cite{AA26}.\\
It is well known that seismic events generate rotational ground motions, which can be effectively measured by RLGs.\\
The integration of rotational and translational measurements improves the understanding of seismic waves, enabling a more refined characterization of earthquake parameters, such as source depth and wavefield properties \cite{s21165344, GA3}.\\
A fast response is important for seismology. To improve real-time capabilities, algorithms such as "Single Tone" (ST)\footnote{In this paper, all mentions of the FFT specifically refer to this algorithm}, a LabView algorithm based on FFT \cite{ni_single_tone_measurements}, or specialized NNs have been implemented. These two methods will be described and compared.\\
We have developed a convolutional NN for frequency reconstruction with minimal delay, suitable for earthquake detection, extracting the frequency from beat notes lasting one-hundredth of a second (50 points at 5 kHz) \cite{DISOMMA2024169758}. Compared with ST, our NN is twice as precise with the same processing delay.\\
The second issue concerns data selection. We have developed an algorithm capable of recognizing various phenomena, such as laser transients and typical laser disturbances. These capabilities are enabled by the automated labeling of data and the implementation of a mask that facilitates the selection of high-quality data.\\
The last issue is to provide a large set of data for earthquake simulation purposes. We have developed a second NN designed to map earthquake signals detected by the "Gruppo Italiano di Geologia Strutturale" (GIGS) seismometers into the signals observed by the co-located GINGERINO ring laser prototype. This mapping is useful for studying rotational models of earthquakes.\\

\section{Data delay line for seismology, triggers and feedback controls}

We have developed an NN capable of extracting the frequency from a window of one-hundredth of a second, which corresponds to a 50-point sinusoid with frequency range between 100 Hz and 500 Hz and compared this NN with ST.\\
We used synthetic sinusoidal signals with different frequencies and phases; we also varied their amplitude and mean values, adding Gaussian noise to avoid over-fitting. Since the real signal always has the same average frequency, the network tends to memorize this value instead of learning to generalize across the wider input window provided by the simulated data, thus losing robustness. Additionally, the network that achieved the best results was the one that outputs both the frequency and the cleaned sinusoid. This strategy ensures that by reconstructing both the clean sinusoid and the frequency, the network learns to better correlate the two pieces of information. \\
In the following, we will go into the details of the construction of this NN, starting with the creation of the dataset, then moving on to the structure we used, and ending with the fit of the hyperparameters to minimize the loss.

\subsection{Creation of a dataset}

\begin{figure}[ht]
\hspace{-10mm}
	\includegraphics[width=0.59\textwidth]{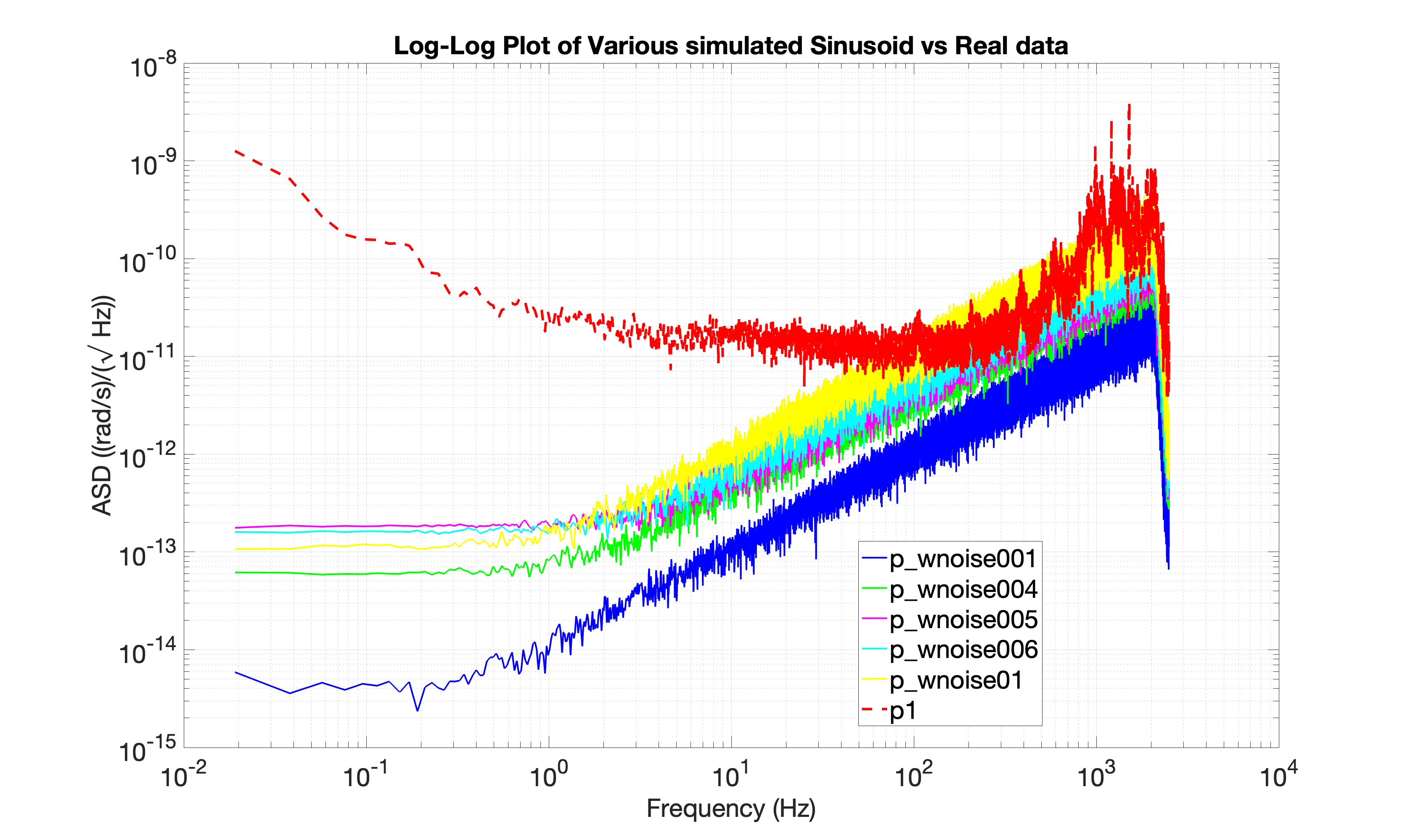}
	\caption{Plot showing the behavior of simulated Gaussian noise compared to GINGERINO data (red) in terms of Amplitude Spectral Density. The red curve represents an estimate of the noise from GINGERINO data, while the cyan curve (“p\_wnoise006”) corresponds to Gaussian noise with $\sigma = 0.006$, providing the best fit to the real data. The agreement is particularly evident around the GINGERINO operating frequency, near 280 Hz, which represents the main band of interest for the experiment.}
	\label{taraturaNoise}
\end{figure}

Creating a dataset for training a NN is a critical step in developing a robust and effective model. The representativeness of the dataset, in how well it reflects real data and effectively simulates noise and potential distortions, directly influences the model's performance and its ability to generalize to unseen data. \\
An initial approach was to feed the network with real data as input and use the frequencies obtained via the Hilbert transform method as output. Although this method was simple to implement, it did not produce satisfactory results because the networks tended to memorize certain features and patterns missing the ability to generalize. This problem arose because all the sinusoids were centered around the operating frequencies of the apparatus (280 Hz for GINGERINO and 184 Hz for GP2, another RLG prototype \cite{AA10}), which led to a loss of robustness. Additionally, the network could not distinguish between signal and noise. For this reason, we decided to change our approach and simulate sinusoids with added noise. However, it is important to note that the noise characteristics of our apparatus are currently under investigation, and the noise used in our simulations is an approximation of white noise generated through a Gaussian distribution.\\
Therefore, we need to simulate a 50-point sinusoid to represent a beat note sampled at 5 kHz over one hundredth of a second, Given that our frequency of interest is approximately 280 Hz, we initially considered simulating sinusoids ranging from 100 Hz to 500 Hz in increments of 0.0001 Hz to train a NN with high accuracy. However, due to memory limitations and the necessity to simulate additional parameter variations to enhance the robustness of our results, we opted to generate frequencies randomly within this range. This approach allowed us to increase the number of sinusoids generated while staying within computational constraints because the size of the dataset plays a significant role in model performance. \\
Another technique used to increase the robustness of the NN is data augmentation, a technique used to artificially increase the size and diversity of the training dataset by applying various transformations, such as rotation, scaling, and flipping the existing data \cite{shorten2019augmentation}. In our case, the adjustable parameters include the sinusoid's peak-to-peak amplitude, initial phase, the number of frequencies used during training, and the noise added to both amplitude and phase.\\
Incorporating a small amount of controlled noise or imperfections into the dataset can be beneficial, as it forces the model to learn more robust features. We have introduced Gaussian noise into their amplitude to model the inherent noise present in real sinusoids. Additionally, we varied the initial phase, randomly added an offset to the mean of the sinusoid, and simulated a random trend in amplitude. These modifications simulate the imperfections present in real data, enhancing the NN's robustness and its ability to generalize to unseen data.\\
For each example, a sinusoidal signal is generated with a frequency $f$ randomly chosen between 100 Hz and 500 Hz:
\begin{equation}
f \sim \mathcal{U}(100\,\text{Hz},\,500\,\text{Hz})
\end{equation}
An offset is added, also randomly selected within a range:
\begin{equation}
\text{offset} \sim \mathcal{U}(-0.2, 0.2)
\end{equation}
to simulate a jitter of the average of the sinusoid points.\\
The initial phase $\phi$ is uniformly distributed between $-\pi$ and $\pi$:
\begin{equation}
\phi \sim \mathcal{U}(-\pi, \pi)
\end{equation}
Initially, phases were varied by fixed increments, but memory constraints limited the achievable coverage. Tests indicated that restricting the phase range led to poor network generalization and reduced frequency extraction accuracy. Thus, we switched to randomly varying the phase within the full $(-\pi, \pi)$ range to better span the parameter space.\\
The crucial aspect of sinusoid simulation to make it as close as possible to the real signal is adding noise to both the amplitude and phase. We started with white noise, modeled as Gaussian noise, and conducted several tests to determine the appropriate $\sigma$ amplitude for this Gaussian distribution. We generated various sinusoids with added Gaussian noise and compared them to real data using the power spectrum. We found that, at high frequencies, where amplitude and phase noise are most significant, the Gaussian noise that best matches the real data has a $\sigma$ of 0.006, see Figure \ref{taraturaNoise}. To obtain an estimate of the apparatus noise, we used the fact that the two signals from the beam splitter outputs are in anti-phase: their difference cancels the Sagnac contribution, allowing us to isolate and estimate the residual noise affecting the system \cite{PhysRevLettNoise}. To enhance the robustness of the NN, we decided to expand the range of added noise.
In light of the above, the noise introduced in the signal has a Gaussian distribution with a standard deviation randomly selected between 0.001 and 0.01:
\begin{equation}
\eta_{Gauss} \sim \mathcal{N}(0, \sigma_{Gauss})
\end{equation}
where $ \sigma \sim \mathcal{U}(0.001, 0.01) $.\\
Similarly, the phase noise is Gaussian with:
\begin{equation}
\eta_{\phi} \sim \mathcal{N}(0, \sigma_\phi)
\end{equation}
where $ \sigma_\phi \sim \mathcal{U}(0.001, 0.01) $.\\
To make the NN more robust and try to simulate as many distortions as possible, we also included a linear trend applied to the amplitude:
\begin{equation}
A(t) = \text{linspace}\left(\text{trend\_start}, \text{trend\_end}, 50 \right)
\end{equation}
where $ \text{trend\_start} \sim \mathcal{U}(0.6, 1.2) $ and $ \text{trend\_end} \sim \mathcal{U}(0.6, 1.2) $.\\
So finally, the clean sinusoidal signal is defined as:
\begin{equation}
s_{clean}(t) = \sin(2\pi f t)
\end{equation}
The noisy signal is then given by:
\begin{equation}
s_{noise}(t) = A(t)\cdot \sin\left(2\pi f t + \phi + \eta_{\phi}\right) + \text{offset} + \eta_{Gauss}
\end{equation}
In our dataset, we generated both the sinusoid incorporating all parameter variations, which we refer to as the complex sinusoid, and its unaltered counterpart, called the clean sinusoid. This approach was adopted because, as we will explain in more detail later, we realized it was necessary to provide the NN with both the complex sinusoid and the clean sinusoid, along with the corresponding frequency. This enabled the network to learn the characteristics of the added noise and to identify correlations between them, effectively performing a form of denoising.\\
We would like to emphasize that the incorporation of these variations was a critical aspect of the training of  NN. In addition to the identification of the optimal network architecture, the implementation of variations was essential to obtain the results that will be discussed below. Each of the described complexities was introduced individually, with careful evaluation of the performance obtained. Each variation was validated separately.\\
The data simulation was conducted simultaneously with the creation of an NN suitable for deriving the frequency; its structure will be introduced in the next section.\\

\subsection{Choice of NN supporting structure}

When selecting a NN architecture to determine the frequency of a sinusoidal signal from 50 data points, it is essential to consider both the physical nature of the problem and the characteristics of the data. The task involves extracting a fundamental physical parameter, like frequency, from a discrete time series, which requires careful consideration of the model's ability to capture periodicity and temporal dependencies.\\
Given the limited number of points and the periodic nature of the data, simpler architectures such as fully connected feedforward NN may be sufficient for this problem, as they can be trained to identify the underlying frequency through direct pattern recognition. However, for more complex or noisy signals, Convolutional Neural Networks (CNNs), might be advantageous due to their ability to detect localized patterns and features within the data, even in small datasets \cite{lecun1998gradient}.\\
In cases where the data exhibits more complex temporal structures or when the signal is embedded in noise, Recurrent Neural Networks, particularly the more advanced Long Short-Term Memory (LSTM) networks, can be employed using an Encoder-Decoder architecture \cite{hochreiter1997lstm}, a widely used framework for sequence learning tasks \cite{sutskever2014seq2seq}.\\
In light of the above, we started with two of the simplest networks to extract the frequency from a 50-point sinusoid: a convolutional network that takes a 50-point input and outputs a single value using a regression method by adjusting the number of neurons per layer, and a "seq2seq" architecture, commonly used for sequence prediction tasks, consisting of an encoder layer to process the input sequence, a bottleneck layer to compress the information, and a decoder layer to reconstruct the sequence. This is followed by a flattened layer, which converts the multidimensional data into a one-dimensional vector, and a dense layer, which acts as a fully connected layer to produce a single-valued output. From these initial models, we experimented by varying their depth and adjusting various parameters, which will be detailed in the training section.\\
None of the networks constructed in this way were able to achieve satisfactory results, as evidenced by the validation metrics and loss values. Validation refers to evaluating the model on a separate dataset that is not used during training to assess its performance and generalization ability. Loss, on the other hand, quantifies the difference between the model's predictions and the actual values. Both metrics are crucial for determining how well the model is learning and performing. Despite making the networks deeper or more complex, which significantly increased the training time, there were no improvements in these metrics, indicating no benefit in the results.\\
The key decision in choosing the optimal layer structure for our specific problem was to establish a connection between the noisy sinusoid, the clean sinusoid, and the frequency derived from them. Initially, we considered including both noisy and clean sinusoids in the training data, but we later decided to have the NN perform a denoising procedure on its own. This means feeding the noisy sinusoid as input and training the network to produce the clean sinusoid and its corresponding frequency as outputs. By doing so, the network can autonomously learn the correlations between the noisy sinusoid and the frequency during training, facilitated by an intermediate denoising step.\\
We found that the most efficient way to construct such a network is to concatenate two CNNs. The first CNN handles the denoising process, acting like a standard 'seq2seq' model, transforming a 50-point noisy sinusoid input into a 50-point clean sinusoid output. This clean sinusoid then serves as the input for the second CNN, which performs a regression process to learn the frequency and outputs it as the final result. In the initial denoising stage, we experimented with using LSTMs, which are typically well-suited for these tasks. However, achieving comparable results to the aforementioned configuration required extensive fine-tuning, which significantly increased the network's complexity. As a result, the LabView program supporting the network struggled to keep up with the incoming data, leading to delays, memory overflow, and eventual program crashes. Therefore, the concatenated double-CNN network, being equally effective and computationally lighter, is the ideal choice for our purposes. \\
\begin{figure}[ht]
 \centering
    \includegraphics[scale=0.45]{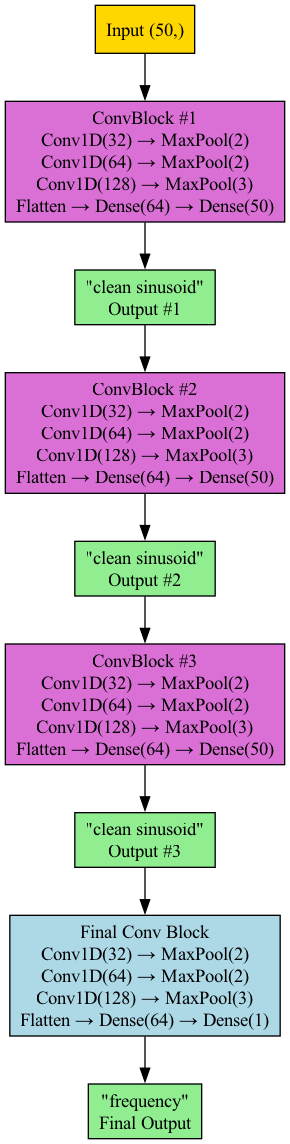}
    \caption{Network structure for frequency reconstruction. The sinusoidal input is processed through three ConvBlocks that perform denoising, producing cleaned sinusoids. The final part of the network estimates the associated frequency, with a dense output layer.}
    \label{ReteFINALE}
\end{figure}
The proposed network architecture, see Figure \ref{ReteFINALE}, consists of a series of CNNs designed to handle both denoising and frequency estimation from a 50-point sinusoidal input.\\ 
Initial attempts to incorporate LSTMs for denoising were discarded due to their higher computational demands and inefficiency in real-time processing, leading to significant delays and memory issues. Instead, the final design employs a series of CNN layers for both denoising and frequency extraction, allowing for a lighter and more efficient model. Each stage of the network underwent rigorous hyperparameter tuning to ensure optimal performance, focusing on maximizing accuracy and minimizing computational costs. \\
The choice of loss function depends on the task formulation. For example, if the task is framed as a regression problem, Mean Squared Error (MSE) is typically used. This is the metric used in our case to evaluate the loss and the validation loss. During the training phase, the network aimed to minimize the validation loss.\\
This iterative process of fine-tuning was essential to achieve a robust and reliable architecture capable of accurately extracting frequencies from noisy sinusoidal data.

\subsection{Discussion of results}

\begin{figure}[ht]
\centering 
 \includegraphics[scale=0.075]{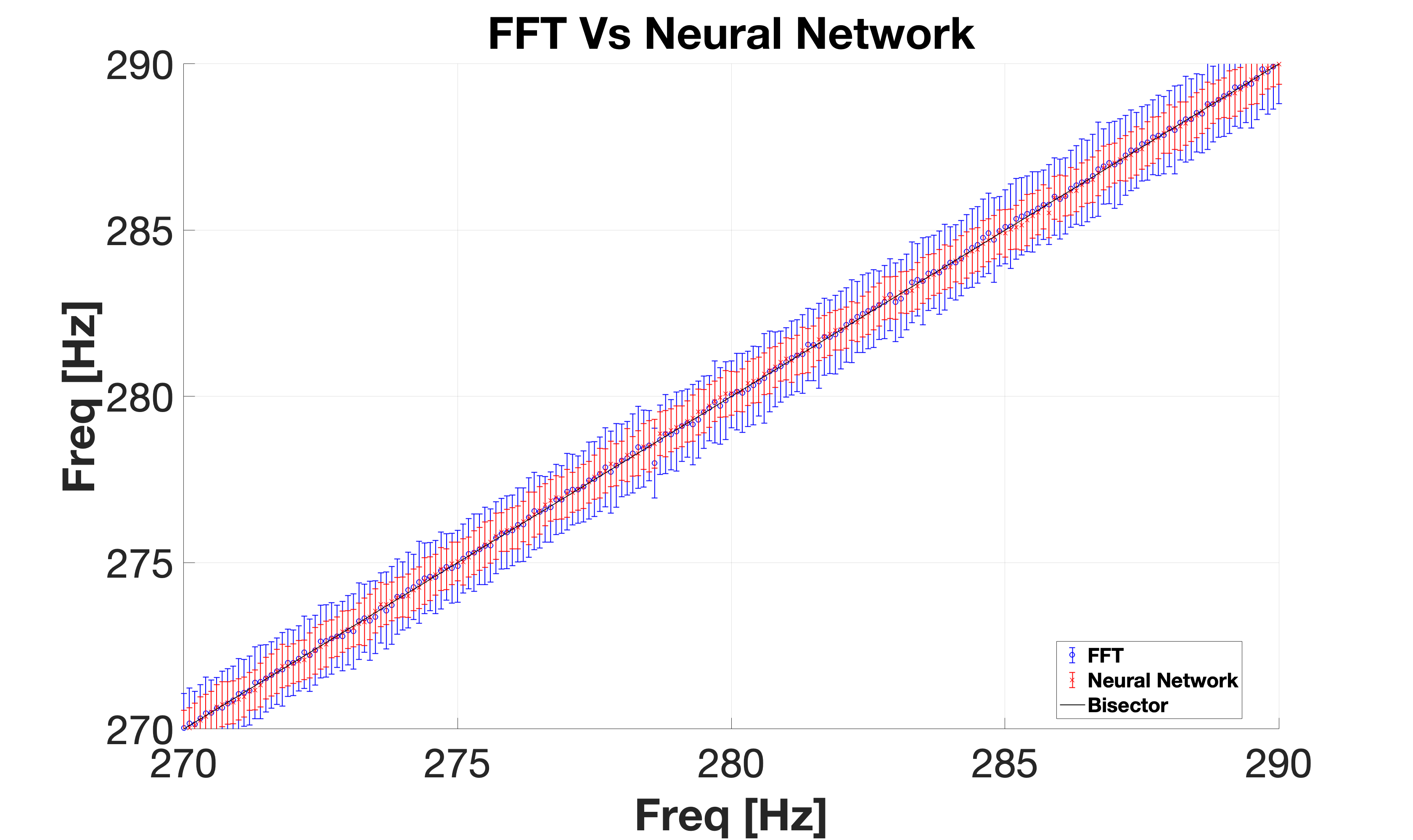}
 \caption{Both the NN and the ST accurately reconstruct the frequency of the simulated signal over the entire frequency range analyzed, but the NN has a spread on average of $1.4$ Hz versus that of the ST of $2.4$ Hz}
 \label{FR1}
\end{figure}

\begin{figure}[ht]

		\includegraphics[scale=0.07]{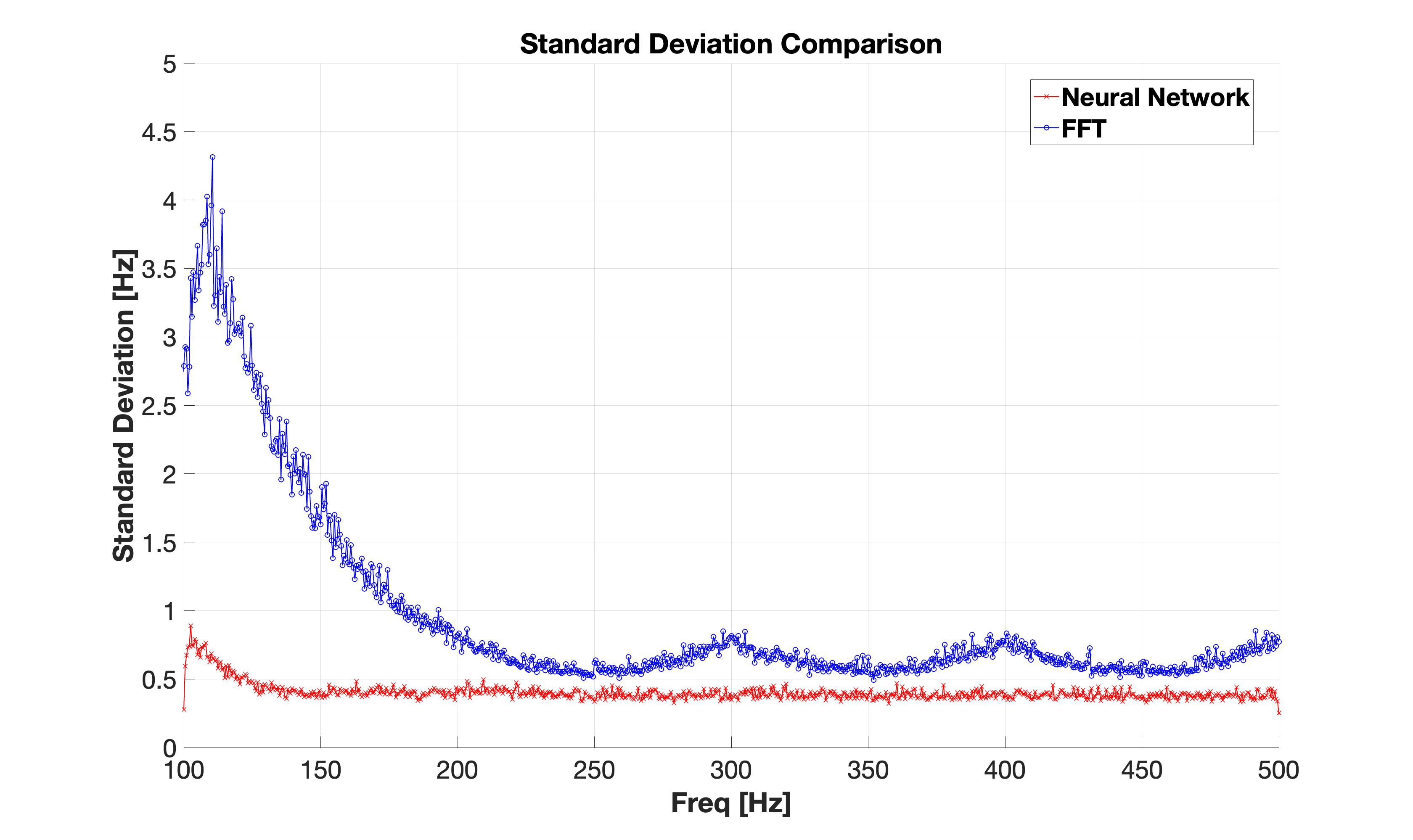}

		\includegraphics[scale=0.07]{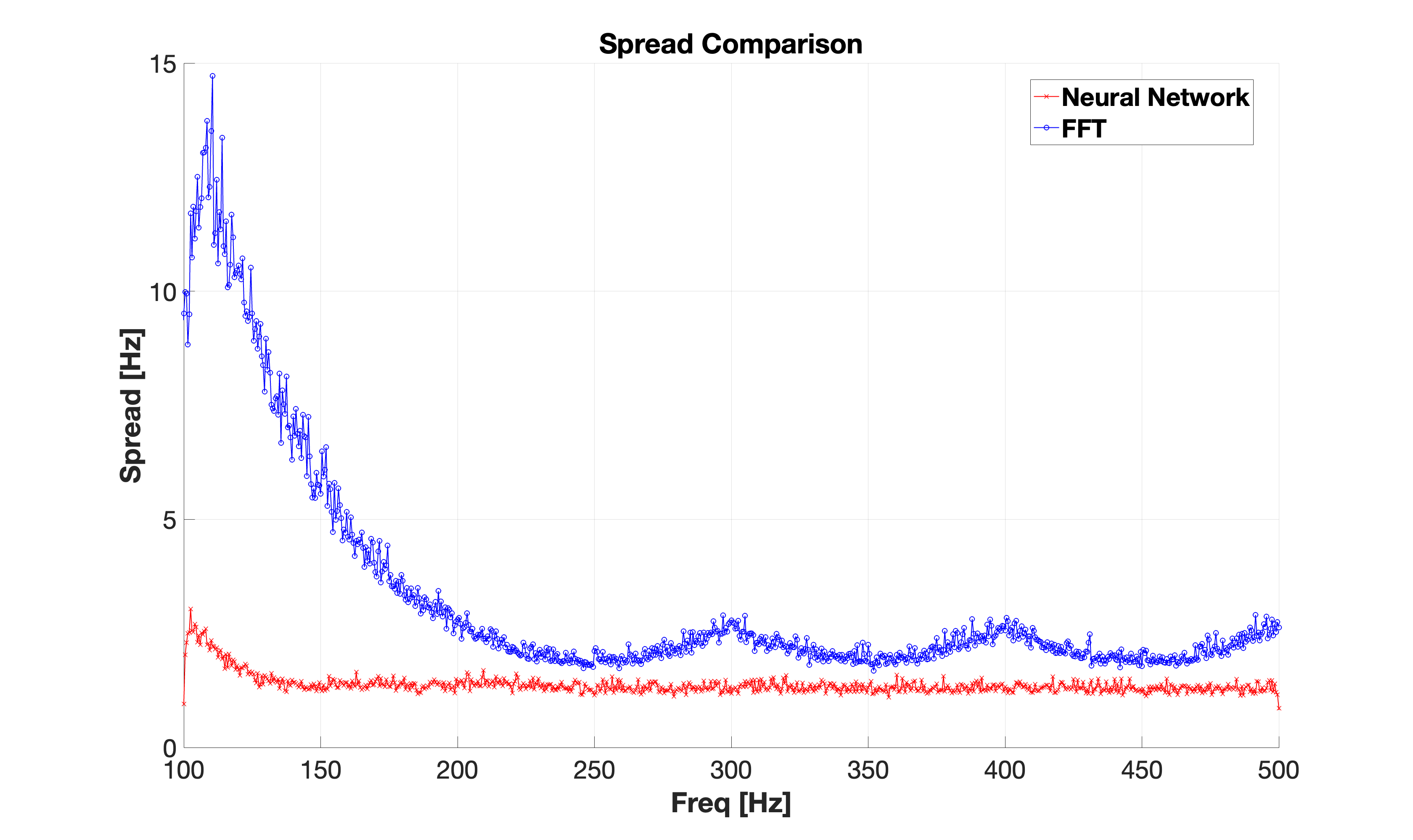}

		\includegraphics[scale=0.07]{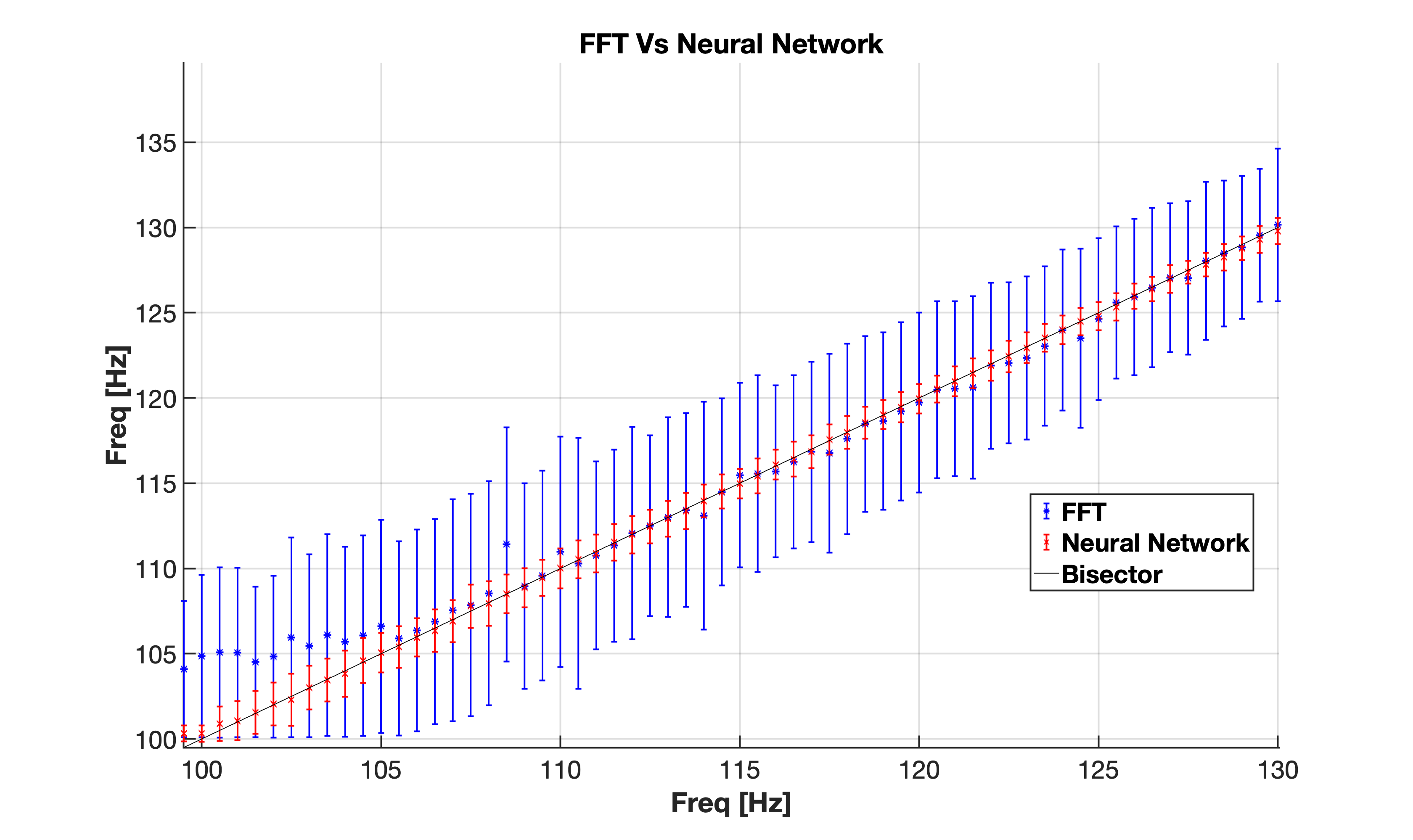}
	\caption{Comparison between the ST and the NN over the maximum training range of the NN. Below 200 Hz, the spread, standard deviation, and accuracy of the ST degrade significantly, while the NN remains more robust and consistent in its response. In fact, the NN outperforms the ST by up to a factor of 4 in the most critical region around 100 Hz.}
	\label{FR4}
\end{figure}

\begin{figure}[ht]
		\centering
		\includegraphics[scale=0.23]{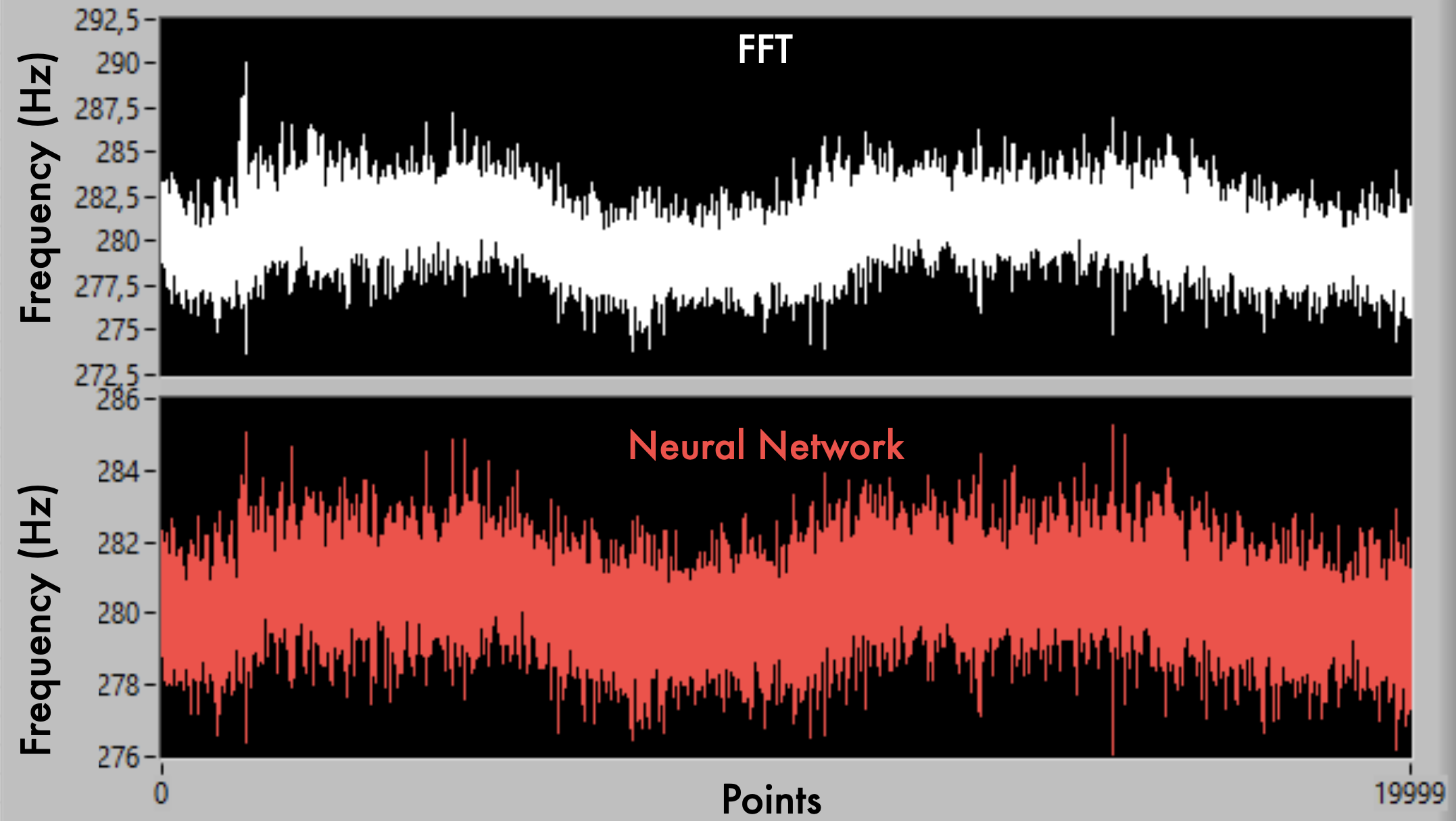}
		\caption{Comparison based on 200 seconds of real data from GINGERINO. On real data as well, the NN outperforms the ST method, showing a lower $\sigma$.}
	\label{RealData}
\end{figure}

\begin{figure}[ht]
		\centering
		\includegraphics[scale=0.22]{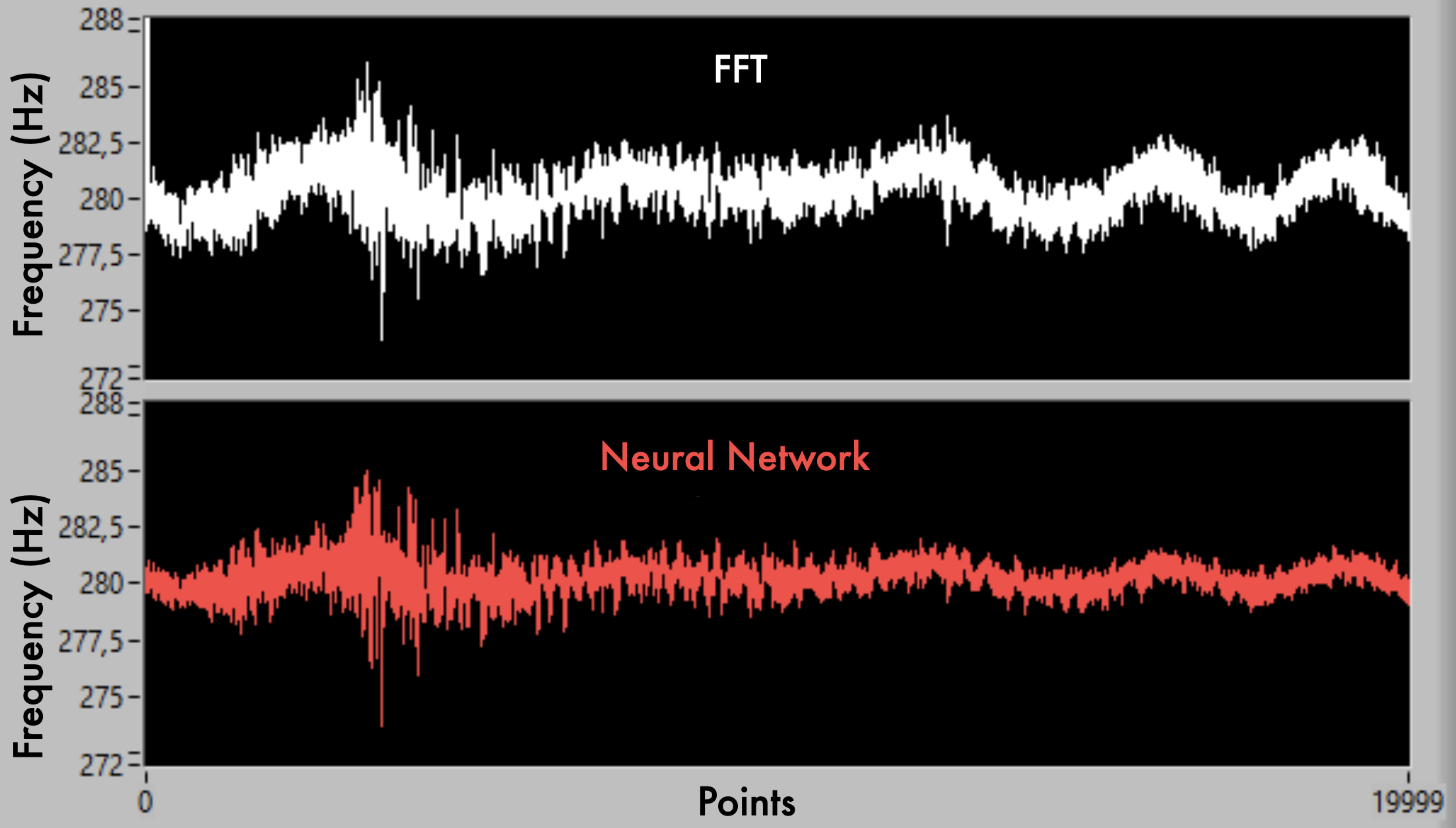}
	\caption{Earthquake detected by both methods. The NN reproduces the same seismic dynamics as the ST. The event took place between November 1 and November 17, 2022, during a period characterized by a seismic swarm.}
	\label{Terremoto}
\end{figure}

Before comparing our two tools for extracting the frequency from a sinusoidal signal using real data (from GP2 and GINGERINO), we need to assess their performance. To do this, we developed a method to obtain frequency distributions and evaluate their statistical properties.
\begin{figure}[ht]
	\centering 
	\includegraphics[scale=0.4]{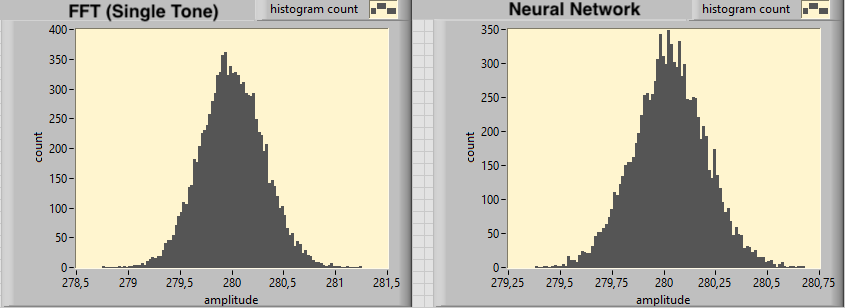}
	\caption{The frequency distributions obtained from 100000 sinusoids with a base frequency of 280 Hz and added Gaussian noise. The distribution on the right is derived from the ST, while the one on the left is obtained from the NN.}
	\label{Gaussian}
\end{figure}
We conducted tests across the range used during the training phase, which includes GINGERINO’s typical operating frequency of 280 Hz, spanning from 100 Hz to 500 Hz in increments of 0.2 Hz. For each target frequency, we generated 100000 simulated sinusoidal signals, adding random noise to mimic measurement uncertainty and real-world imperfections. After generating these noisy signals, we applied the ST and NN to extract the dominant frequency of each signal. The ST decomposes a time-domain signal into its frequency components, but due to the presence of noise and the finite duration of the signals, it does not always yield a precise estimate of the original frequency. The NN also exhibits its frequency dispersion due to limitations in learning the complex signal. However, as we will show, this dispersion is smaller compared to that of the ST. \\
Consequently, we obtained a distribution of estimated frequencies for each target frequency using both methods, as shown in Figure \ref{Gaussian}, these distributions are approximately Gaussian. \\
To analyze them, we calculated two key metrics: the first is the standard deviation of the distribution, assuming it to be Gaussian; the second metric is the spread, defined as the difference between the highest and lowest frequency bins containing output values, providing a measure of the variability in frequency estimation across the two methods.
\begin{figure}[ht]
	\centering 
	\includegraphics[scale=0.075]{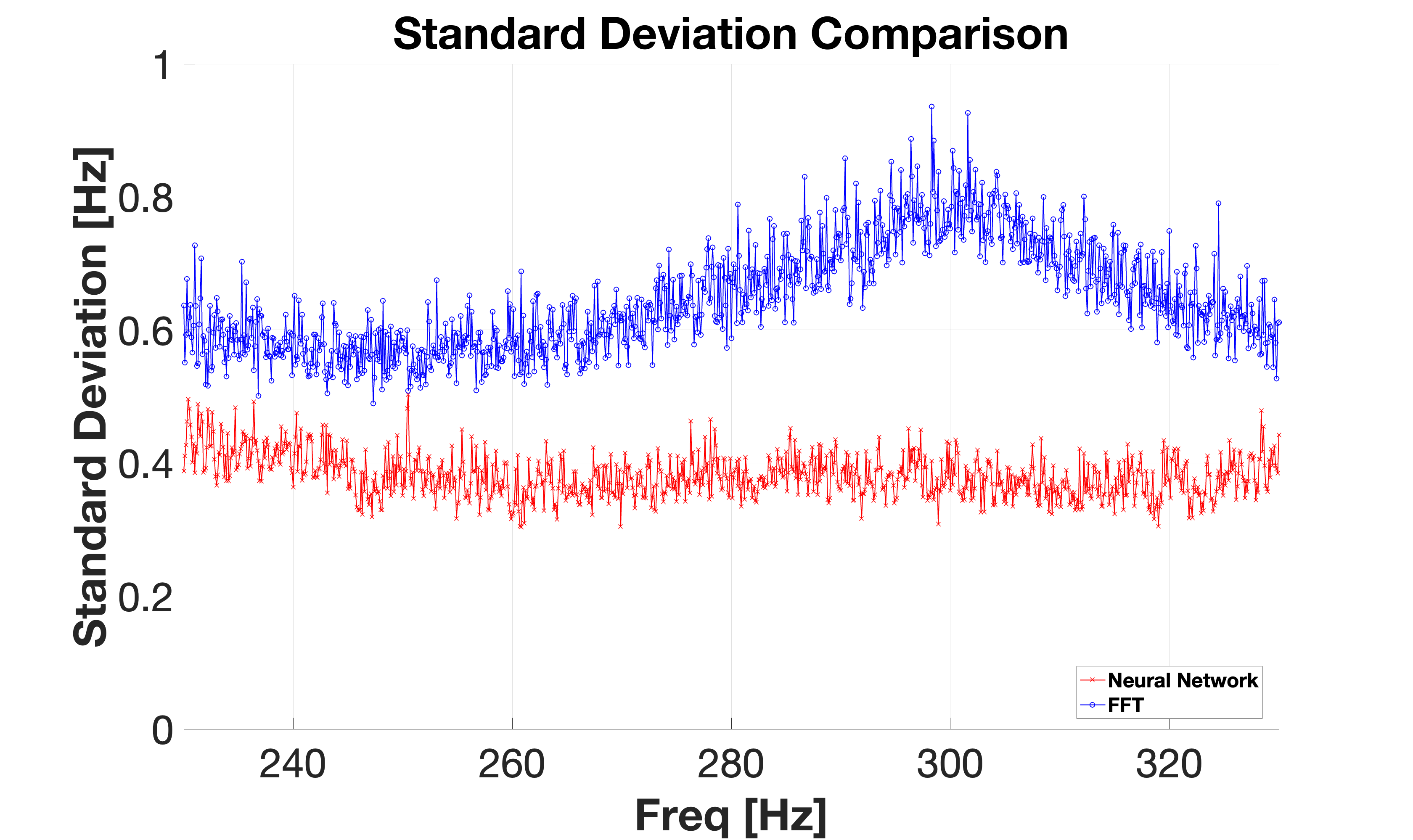}\quad
	\includegraphics[scale=0.075]{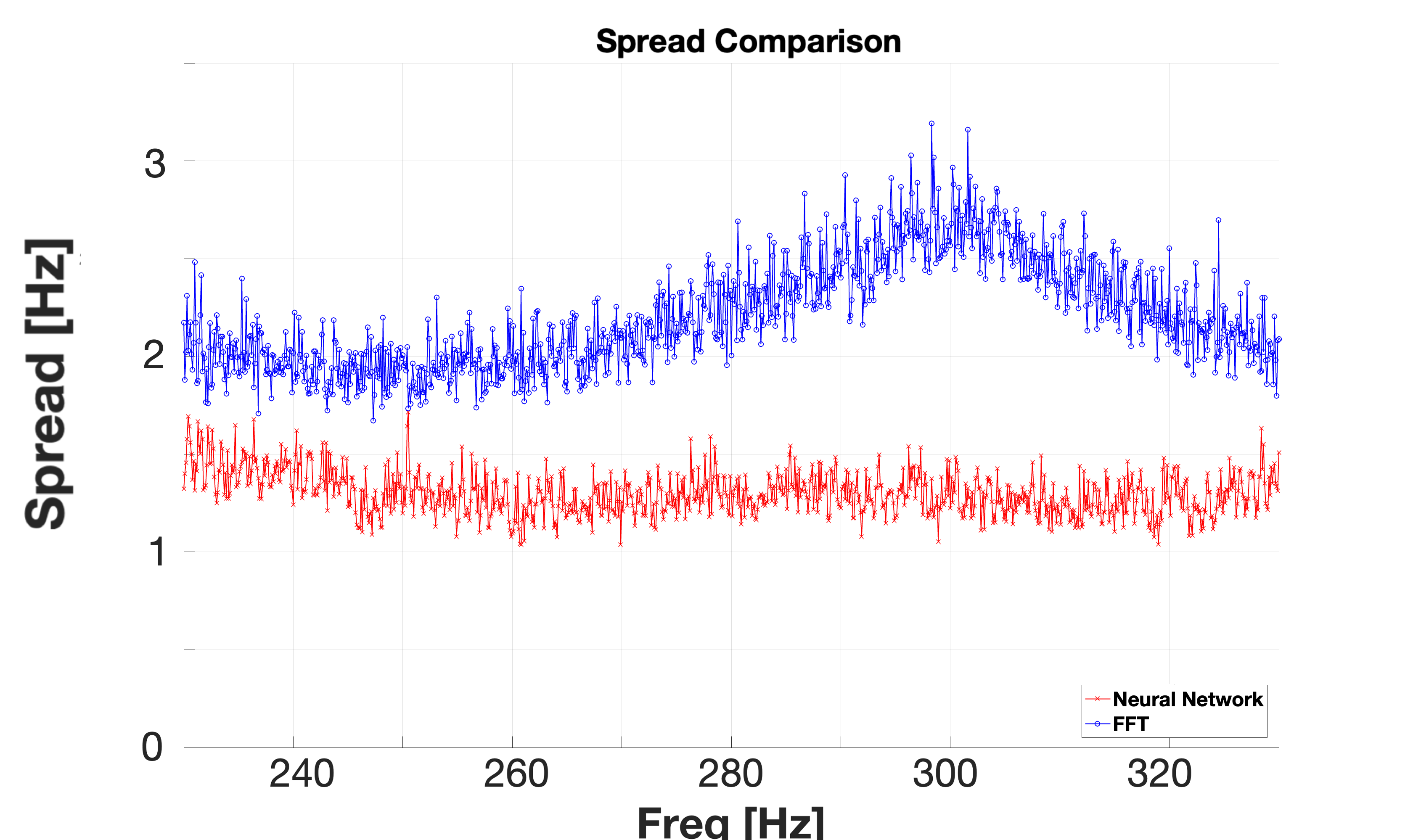}
	\caption{Comparison of the standard deviation and spread of the frequency in the region of GINGERINO’s typical operating frequency, obtained from an NN (in red) and an ST (in blue). The superior precision of the NN across the analyzed frequency range is evident from both metrics.}
	\label{FR2}
\end{figure}
As shown in Figure \ref{FR2}, for both metrics, standard deviation and spread, the NN outperforms the ST, with an average improvement of a factor of 2. Another advantage of the NN is the absence of significant fluctuations across the entire analyzed range, resulting in consistent metrics and making the network more robust. We will examine in detail what happens to the ST performance over broader ranges in the appendix \ref{A1}. \\
Focusing on the accuracy of these two methods within the frequency range very close to GINGERINO’s mean operating value, both demonstrate high precision, with the expected value falling within the obtained distributions, even within their respective standard deviations, see Figure \ref{FR1}. However, the situation changes when we expand the range. 
Repeating the test over a broader range, from 100 Hz to 500 Hz, the maximum training range of our NN, the results shift further in favor of the NN. In this expanded range, both the standard deviation and the spread for the NN remain stable, except for a slight decline at lower frequencies around 100 Hz. In contrast, the ST shows oscillatory behavior near multiples of 100 Hz and a significant degradation, by an order of magnitude, below 200 Hz. This decline is evident in the standard deviation, spread, and accuracy of the ST, as illustrated in Figure \ref{FR4}. The mean of the ST values deviates considerably from the expected value, remaining within the broadened distribution but failing to fall within the standard deviation for the lower frequency values.
The ability of the NN to ensure a stable and consistent response also provides greater flexibility in tuning specific system parameters. As shown in Equation \ref{eq:SagnacFrequency}, variations in quantities related to the scale factor, such as the area and perimeter of the apparatus, or changes in the gyroscope’s orientation, and thus its projection along the Earth’s rotation axis, affect the corresponding Sagnac frequency. Consequently, if the goal is to shift the operating frequency towards approximately 100 Hz, either by reducing the instrument’s dimensions or by tilting it, employing the NN to deliver measurements with minimal latency constitutes the most robust and reliable solution.\\
After all the tests conducted on simulated data, we now present tests performed on real data. As shown in Figure \ref{RealData}, even with real data, the NN exhibits a smaller spread and standard deviation compared to the ST. In this particular case, the ST exhibits a spread of $16.2$ Hz and a $\sigma$ of $4.7$ Hz, while the NN shows a spread of $9.1$ Hz and a $\sigma$ of $2.7$ Hz, which corresponds to approximately the factor of 2 also observed in the previous tests on synthetic sinusoids, in the region of GINGERINO’s mean Sagnac frequency.
Since the initial layers of the network were designed to act as filters, we investigated whether the network would still be capable of preserving the full dynamics of a seismic event when applied to real data. Therefore, we searched for real seismic events within the data and selected a representative earthquake signal to test our frequency reconstruction methods, verifying their ability to track its profile without distortion, as shown in Figure \ref{Terremoto}. In this case, the NN demonstrated a superior signal-to-noise ratio in detecting the earthquake. Moreover, the NN faithfully reproduces the dynamics of the earthquake, comparable to the performance of the ST algorithm.

\section{Applications: classification of laser disturbances and seismic events}

\subsection{Mask for Real-Time Data Classification}

\begin{figure}[ht]
\centering 
 \includegraphics[scale=0.35]{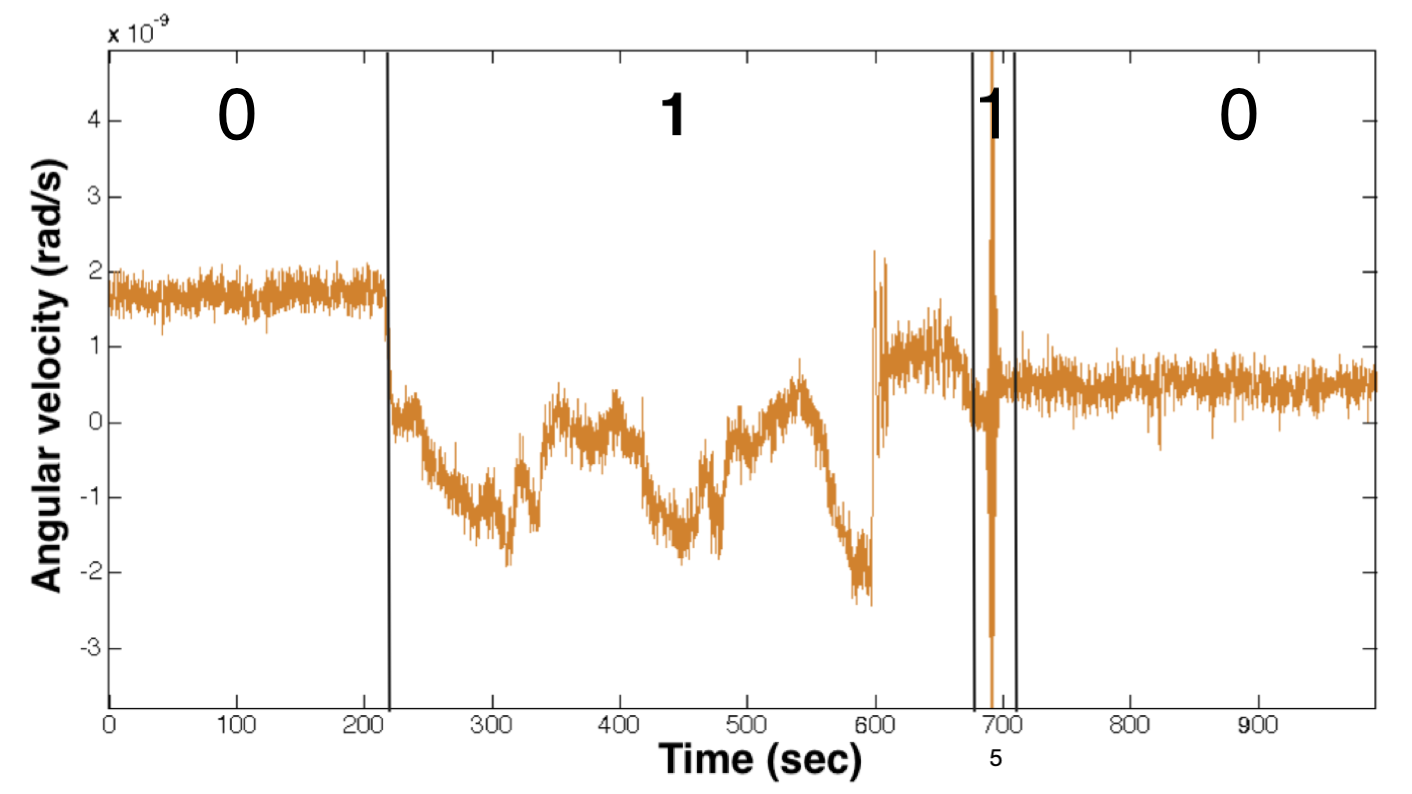}
 \caption{Typical Laser disturbances like mode jump (The smaller portion of the plot around 700 s), preceded by a phase of instability.}
 \label{Spk}
\end{figure}

\begin{figure}
\centering 
 \includegraphics[scale=0.33]{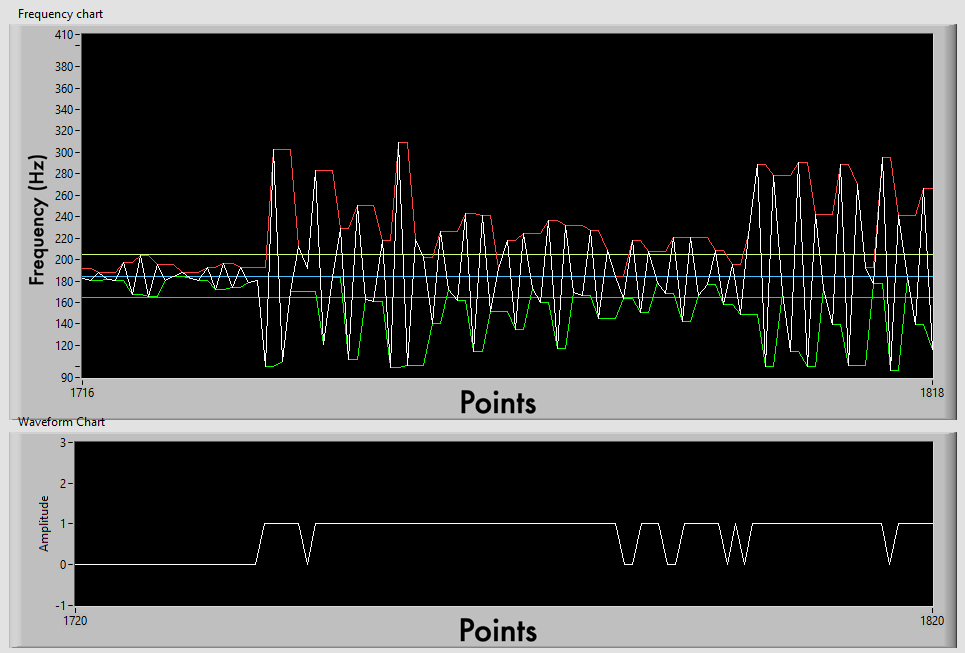}
 \caption{GP2 disturbances. Top: GP2 signal (white) exceeding the $2\sigma$ threshold after a mechanical excitation. Bottom: corresponding real-time mask output.}
 \label{GP2}
\end{figure}

Although GINGERINO analysis methods already have systems to classify the goodness of the signal and currently have a duty cycle of more than 90$\%$, we are building an algorithm that identifies disturbances from the laser. To this end, the input is the GINGERINO signal reconstruction, obtained from a previous NN that contains these disturbances, while the output is one mask that distinguishes between "0", the good signal, and "1" or "2" depending on what the anomaly is, see Figure \ref{Spk}.\\
When the signal is “good", meaning it falls within a frequency range that remains within two $\sigma$ of the disturbance-free signal. To accomplish this, we utilized the signal envelope to efficiently determine when the signal deviates from the previously established frequency band, see Figure \ref{GP2}.
A value of "1" is assigned by the mask when the signal falls outside this frequency band, which typically occurs during brief transients, earthquakes, or when the system is unstable, but not sufficiently to enter split mode: a regime in which the two counter-propagating beams of the ring laser oscillate on different modes, producing a frequency separation equal to the cavity’s free spectral range. In this condition, the beat-note frequency becomes too high to be detected by the photodiode. In the case of split-mode, where the measured frequency is unstable and fluctuates randomly within the whole band defined by the working range of the NN, the signal may intermittently fall within the previously defined "good" band despite not representing a true physical signal originating from the Sagnac effect. In such cases, the mask assigns a value of "2" due to the corresponding fringe contrast dropping below 0.5.\\ 
However, both mode jumps (laser disturbances) and earthquakes are classified as event "1" because both can exceed the chosen two-$sigma$ threshold used to trigger the mask. For this reason, we developed another NN capable of distinguishing seismic events within our data, even those that are immersed in noise and thus cataloged as "0" data.

\subsection{Detection of seismic events through Artificial Intelligence}

\begin{figure}[ht]
    \centering
 \includegraphics[scale=0.4]{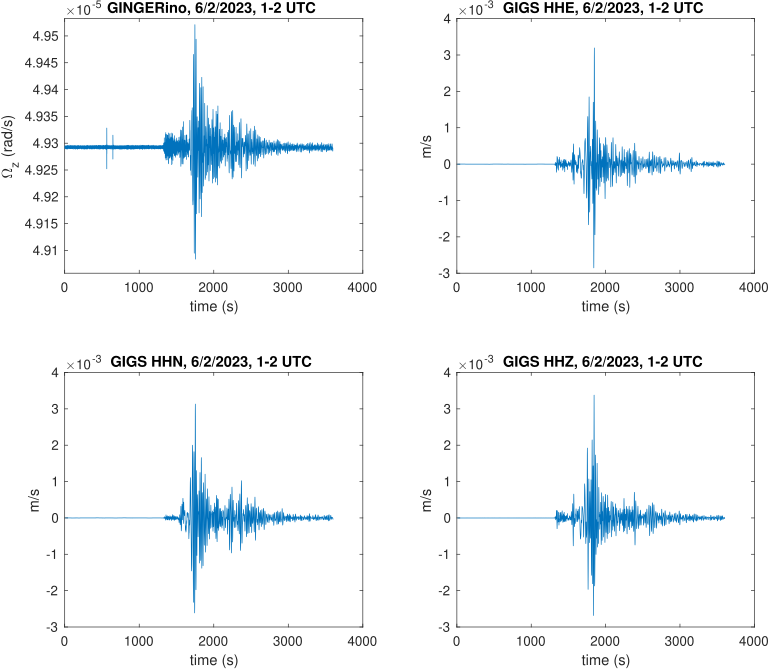}
 \caption{Seismic event seen by GINGERINO (top left), Turkey, February $6^{th}$, 2023 UTC, compared to the same event seen by the co-located seismic station GIGS, East-West component(top right), North-South component (bottom left), z component (bottom right). The mean value in the GINGERINO plot was not subtracted to show the measured Earth angular velocity. Plots show one hour of data, from 1:00 to 2:00.}
 \label{fig:gino_gigs_nofilter}
\end{figure}
\begin{figure}[ht]
    \centering
    \includegraphics[scale=0.4]{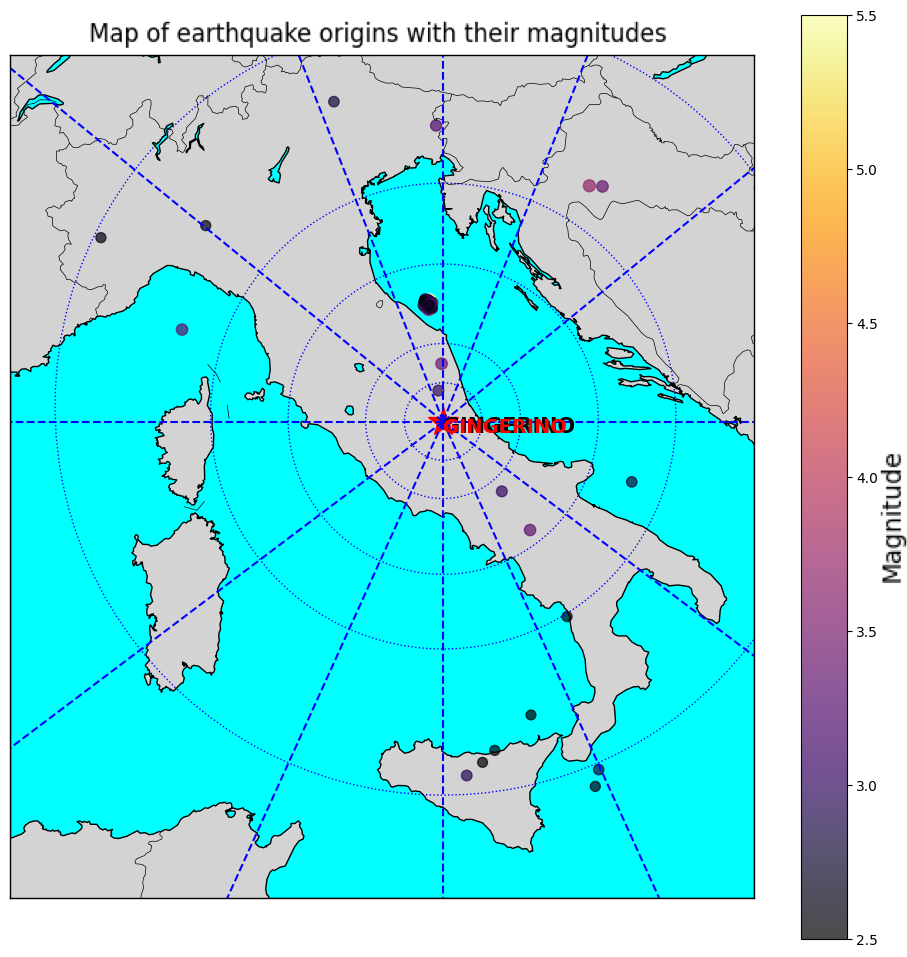}
    \caption{Map of earthquake sources used for training. It should be noted that the highest concentration of events, with 75 located in a single box, occurs slightly north of GINGERINO. Conversely, no events are present within the first 30 degrees of the first quadrant or in the third quadrant, using a Cartesian coordinate system with its origin at GINGERINO.}
    \label{fig:Mappa}
\end{figure}
\begin{figure}[ht]

		\includegraphics[scale=0.32]{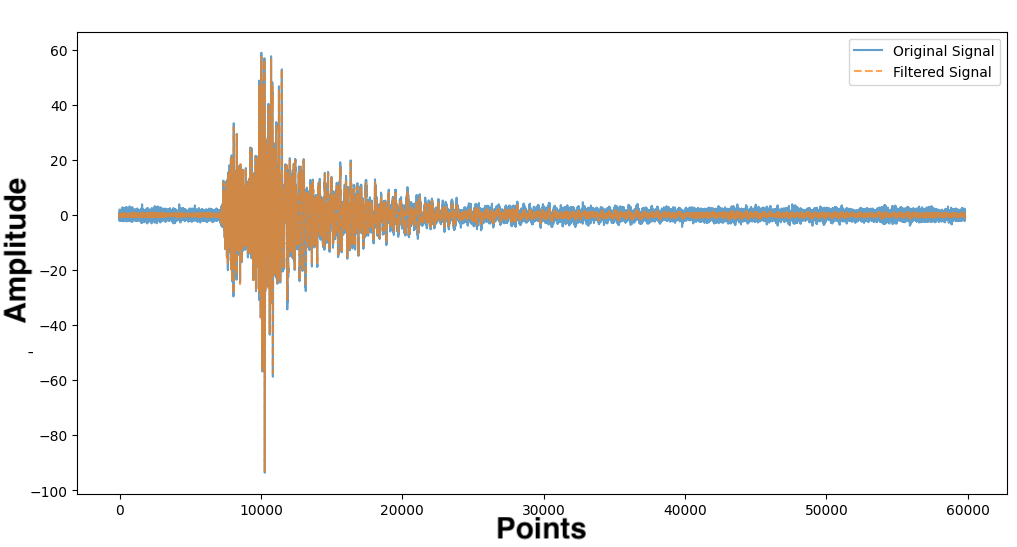}
		\includegraphics[scale=0.32]{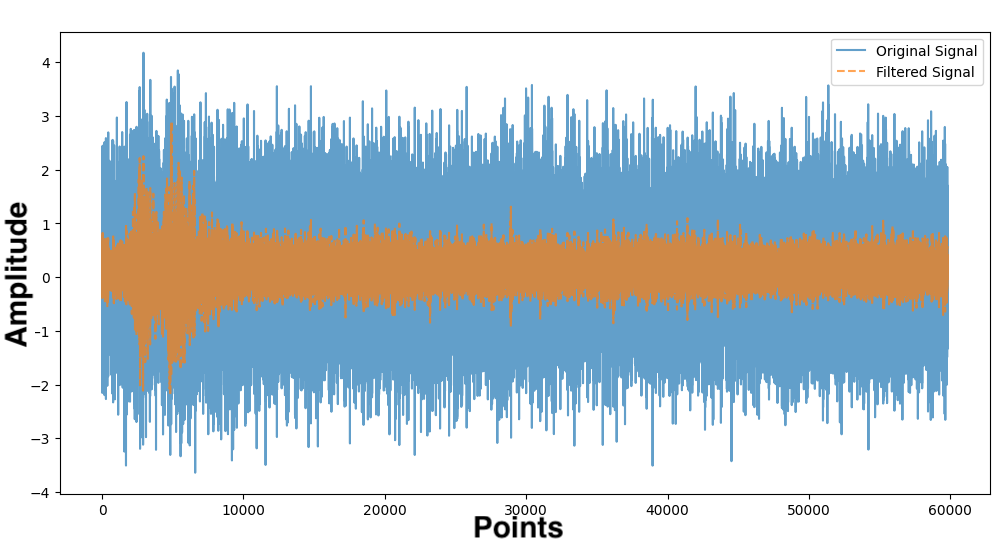}
		\includegraphics[scale=0.32]{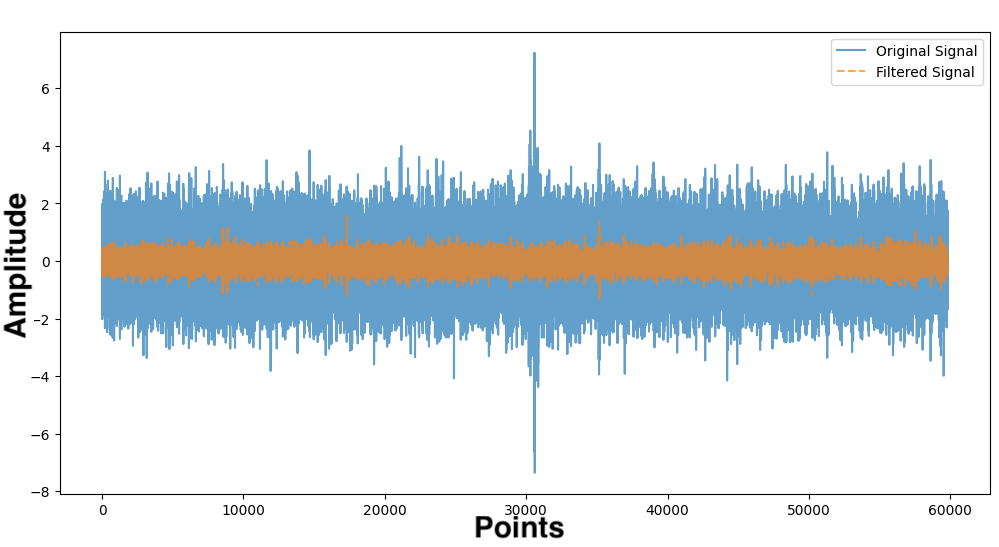}
		\includegraphics[scale=0.32]{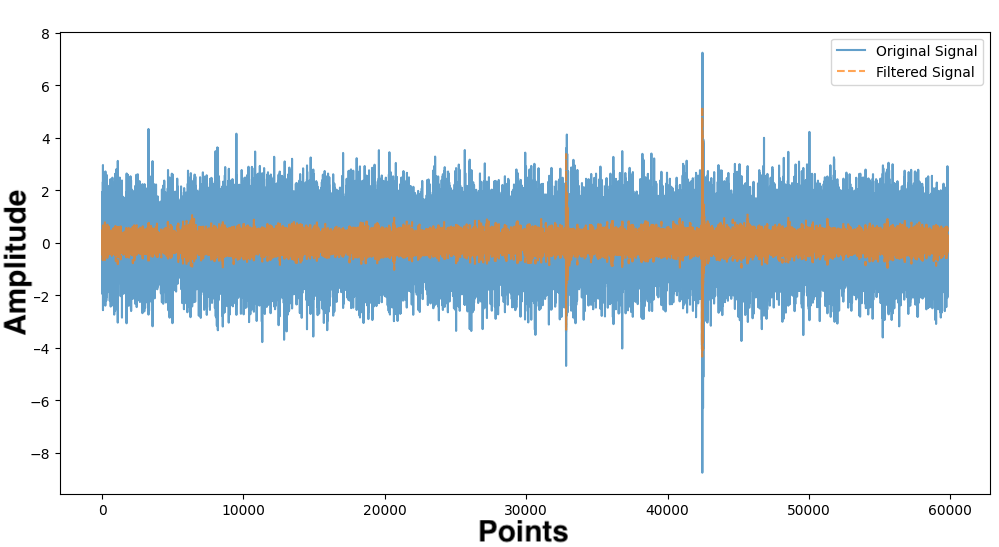}

	\caption{Four examples of earthquakes used for training, in blue the signal we passed to the NN, in yellow it filtered below 20 Hz to highlight the earthquake from the noise. For each event, we have the number of data points corresponding to 10 minutes at a sampling rate of 100 Hz.}
	\label{TP}
\end{figure}
\begin{figure}[ht]
		\includegraphics[scale=0.32]{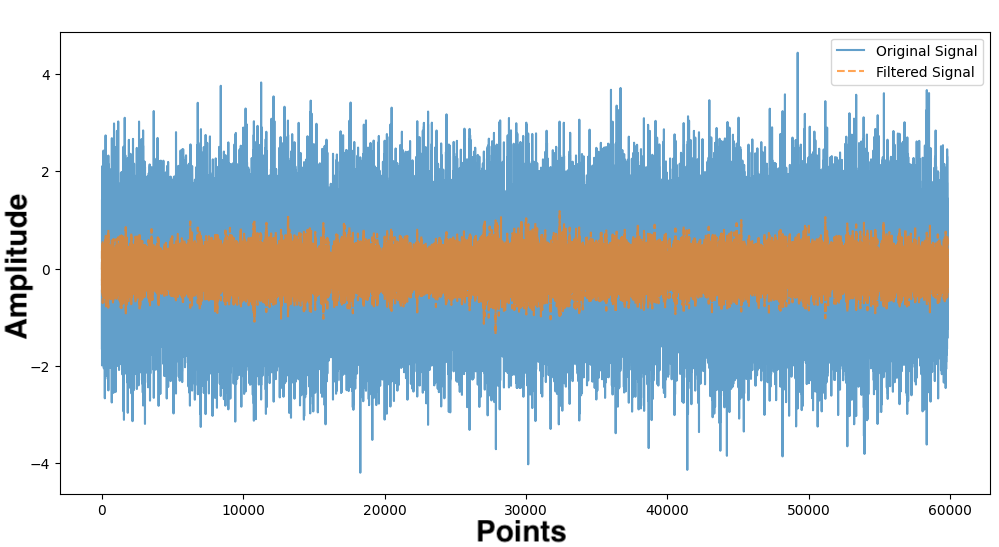}
		\includegraphics[scale=0.32]{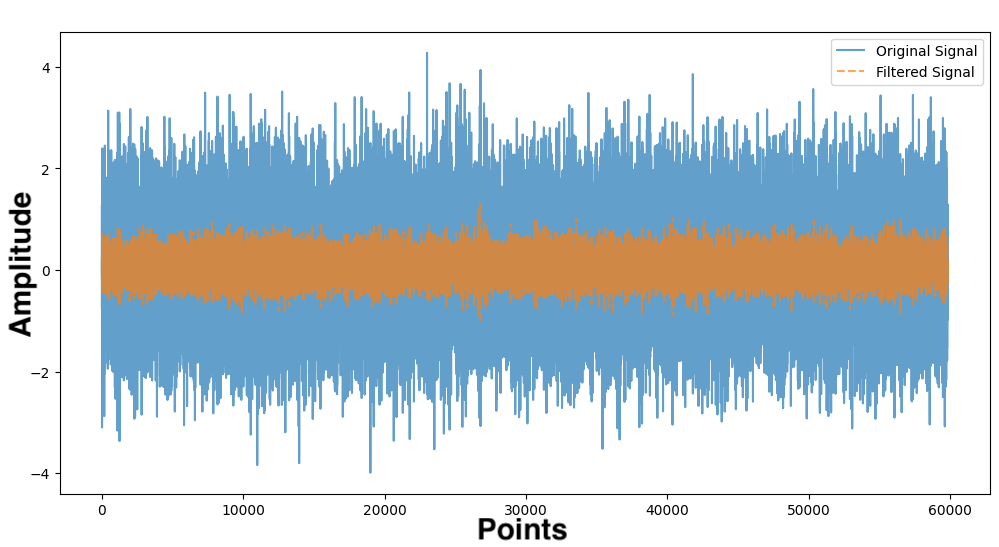}
		\includegraphics[scale=0.32]{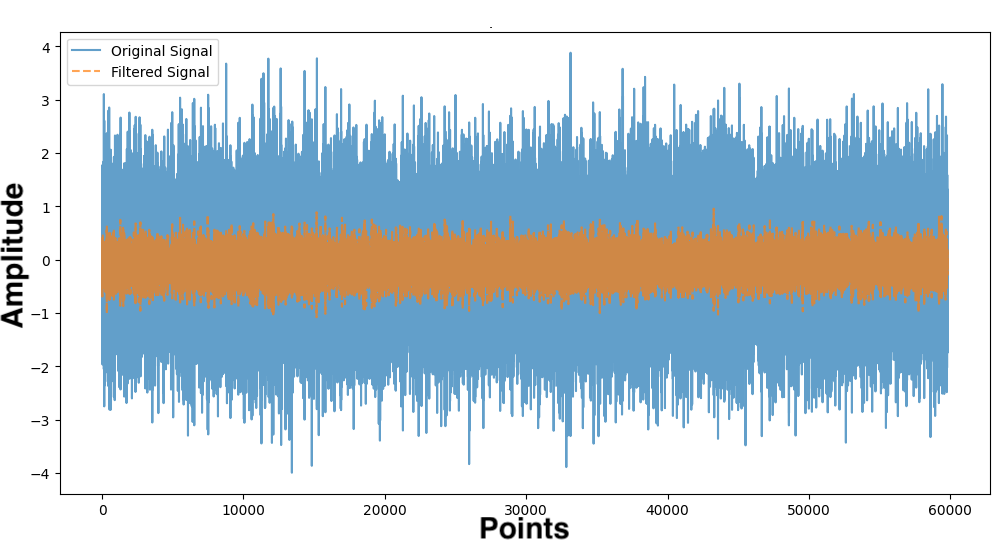}
		\includegraphics[scale=0.32]{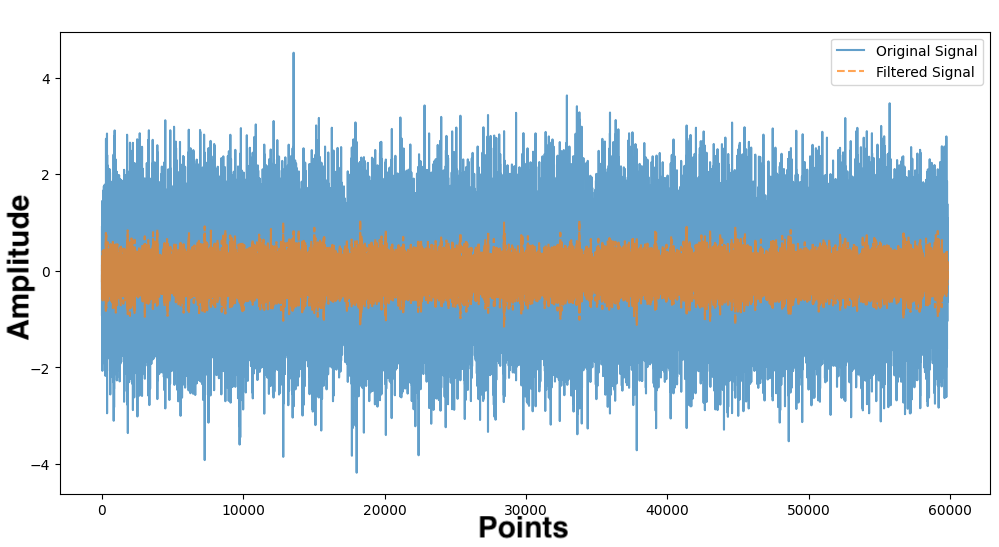}
	\caption{The first two signals contain earthquakes, whereas the last two signals do not. However, even with the filtered signal below 20 Hz (in yellow), the presence of earthquakes is not as evident as in previous cases.}
	\label{FP}
\end{figure}
\begin{figure}[ht]
	\centering 
	\includegraphics[scale=.7]{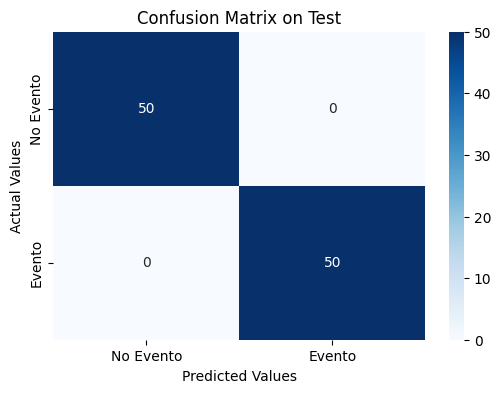}
	\includegraphics[scale=.7]{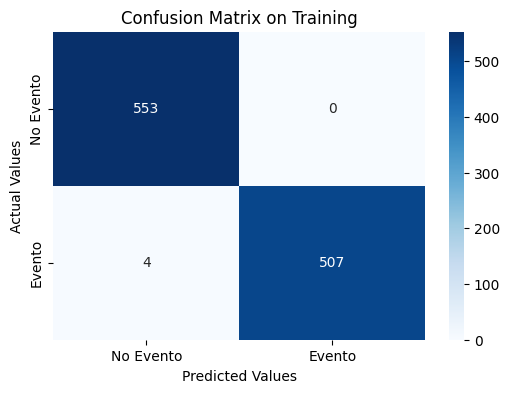}
	\caption{Top: Confusion matrix on test data, unseen during training. Bottom: Confusion matrix on the data used for training.}
	\label{CMatrix}
\end{figure}
\begin{figure}[ht]

		\includegraphics[scale=0.32]{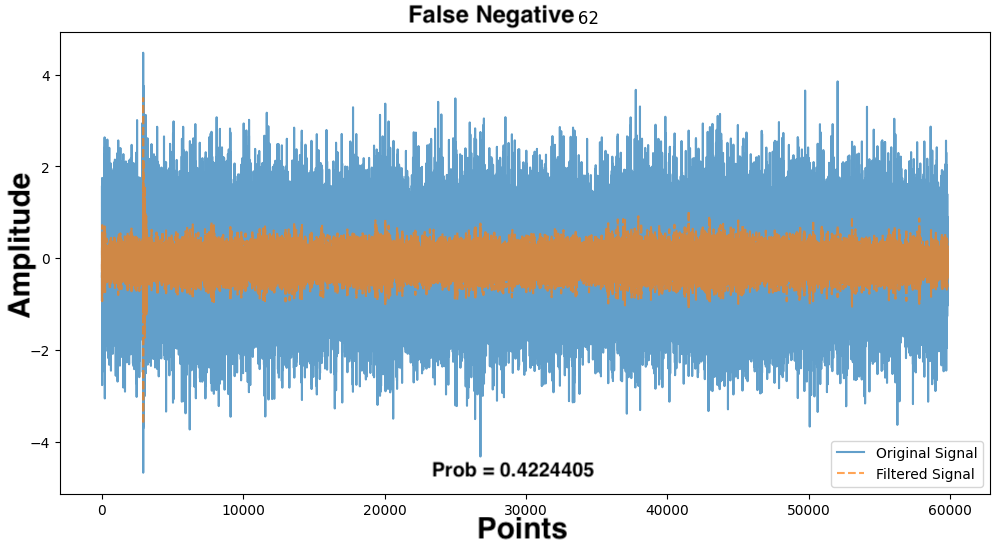}
		\includegraphics[scale=0.32]{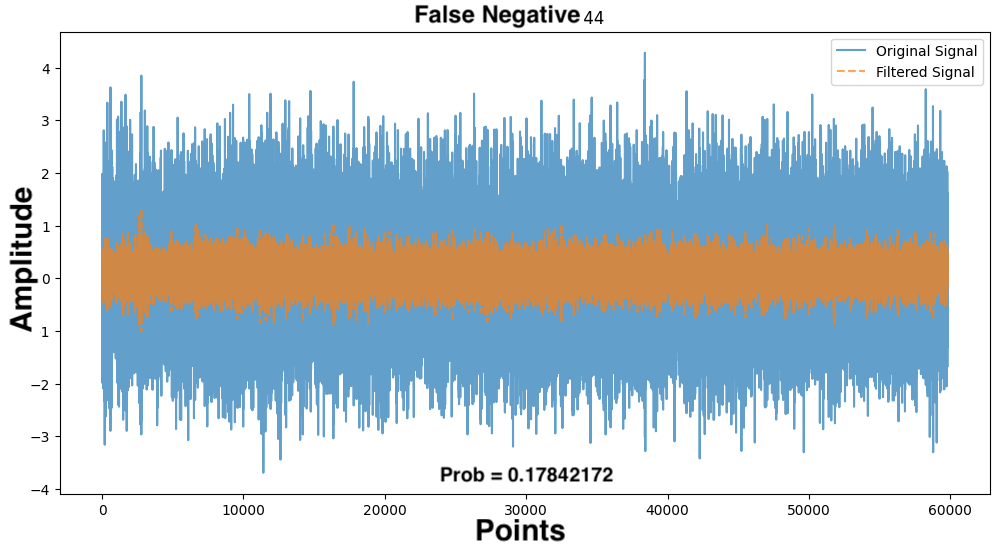}
		\includegraphics[scale=0.32]{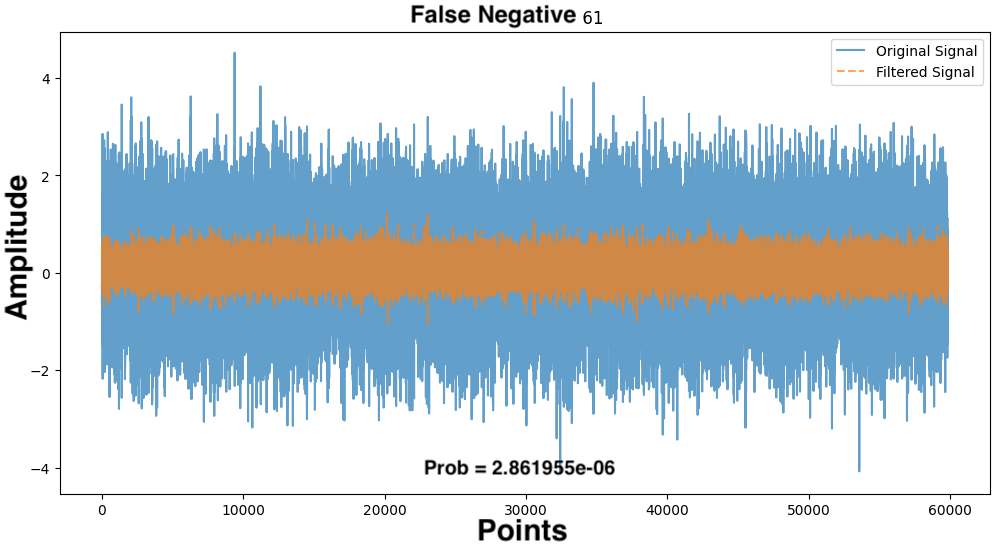}
		\includegraphics[scale=0.32]{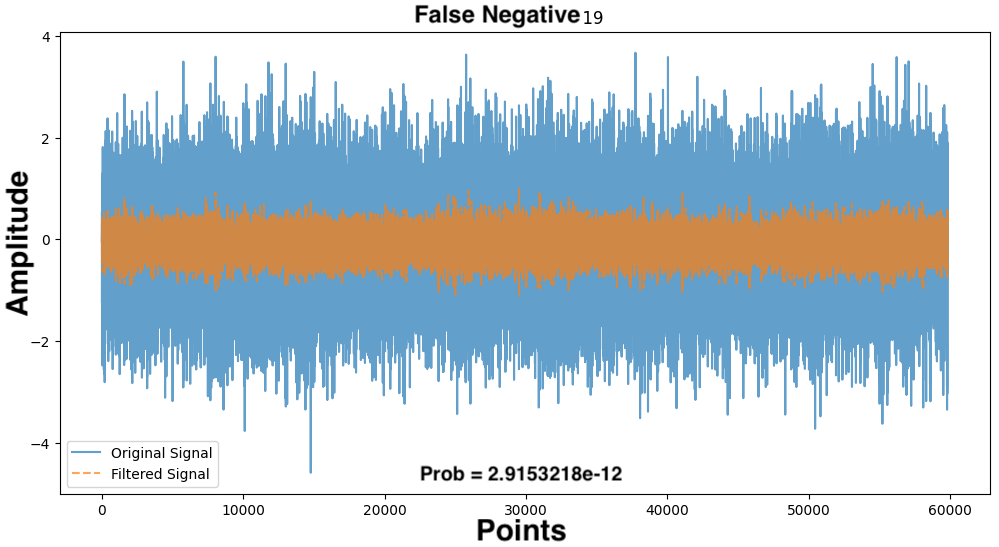}
	\caption{The four events from the training dataset that the network fails to positively identify as earthquakes. It should be noted that, for the first event, the NN assigns an earthquake probability of 42\%, which is close to the 50\% threshold used to classify the presence of an earthquake. This example is also the one where the earthquake is more evident compared to the others.}
	\label{FN}
\end{figure}

The rotational signal of a recent Turkey earthquake seen by GINGERINO, see Figure \ref{fig:gino_gigs_nofilter}, shows that our RLG is at least as sensitive as conventional seismometers currently used for detecting teleseismic events and local earthquakes, particularly with regard to the rotational component of ground motion \cite{lavoro_sismologia}.\\
Therefore, we have developed a NN capable of recognizing seismic events based on GINGERINO data. In addition to improving the classification previously introduced, this NN aims to identify local seismic events that may be obscured by noise and have not been detected by the GIGS seismometer. Once a trigger is launched, we can perform an analysis of these events using both instruments, allowing us to triangulate the data to compare and verify the information obtained. This approach enhances the robustness of our analysis and potentially reveals a range of seismic events that may have gone unidentified.\\
Being able to only simulate teleseismic and retrieving examples of local earthquakes is the main problem. This study is to be considered a first feasibility test, in the future, we could enrich our dataset with new events and integrate them with simulated events through the tools already available in Python, such as "ObsPy" \cite{obspy}. We did not use simulated events for this network; instead, we collected real events from GINGERINO. To do this, we used the INGV catalog of seismic events \cite{Catalogo}, which allows for filtering by location, magnitude, and date (UTC). We selected events in the Italy region with a magnitude greater than 2.5 between November 1 2022, and November 17 2022, a period marked by a seismic swarm. This resulted in 94 seismic events of varying magnitudes and depths, primarily originating from the Marche coast, see Figure \ref{fig:Mappa}. We will revisit this point later when discussing the network's robustness and the tests conducted.\\
Starting from these events, we created 10-minute windows around each event using the available data. We chose this specific duration because it typically includes all local (within 150 km away) and regional (within 500 km away) seismic events, as well as teleseismic events, which have an average duration of 10 to 15 minutes. At our sampling rate of 100 Hz, ten minutes of input corresponds to 60000 data points, which approaches the memory limit for data training that the computer can handle for about a thousand events for each training.\\
To further enhance the dataset, we perform augmentation by generating additional examples from each event through window shifts around the earthquake, positioning it both at the beginning and the end of the record. This approach prevents the network from only seeing centrally located events, thereby improving its robustness. In this way, we obtained examples of seismic events that are labeled as output "1" in our network. To complete the dataset with "0" events (indicating the absence of a seismic event), we searched among the data days in which no catalogued events occurred.\\
Seismic signals are often embedded in noise and require ad hoc filtering to be identified. Our goal, however, is to create a network capable of distinguishing noise from signal without such preprocessing. In Figure \ref{TP}, we show four examples of earthquakes used for training, where the seismic events are evident. 
In contrast, Figure \ref{FP} presents two examples of earthquakes and two non-seismic cases that are difficult to distinguish. It is challenging to determine the presence of an earthquake simply by observing the data, even after filtering below 20 Hz.\\
It should be noted that all the examples have zero mean. This is because, before passing the data to the network, we removed the mean and normalized by their standard deviation to improve the network's learning process.
From these data, we discarded over 300 events, both seismic and non-seismic, due to the presence of laser disturbances or other unidentified noise sources. At this stage, we prioritized simplifying the network's task, with plans to gradually increase complexity in the future by introducing glitches or laser noise that do not correspond to earthquakes. This will allow us to assess whether the network produces false positives when exposed to such disturbances.\\
The total number of events used for this training is 1164, of which 50 "0" events and 50 "1" events were set aside to test the network on samples that were never presented during the learning phase. The remaining 1064 events, consisting of 511 earthquakes and 553 non-events, were split into training and validation data. To maximize the use of all events during training, we applied the folding technique: the dataset (excluding the test set) was divided into four different folds, and the network was trained 4 times, using a different fold each time for validation. At the start of each training with a new validation fold, the parameters were reset to their initial values, while the best-performing network across all folds was saved.\\
In the following sections, we will discuss how the best network was identified, which structure achieved the best results, and the maximum accuracy obtained in this initial phase of training.
\begin{figure}[ht]
\centering
    \includegraphics[scale=0.4]{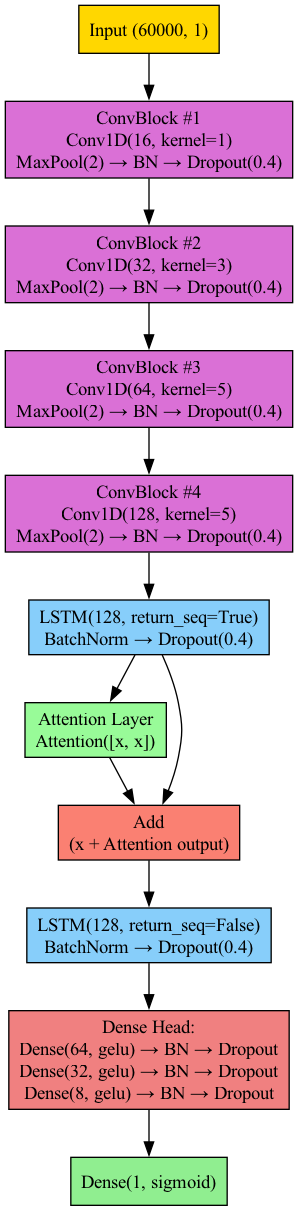}
	\caption{Block diagram of the structure of the network used for classification. After the convolutional layers, which extract the most important features, the LSTM layers process these features by analyzing temporal dependencies. Attention layers are then applied to identify the most relevant features, and the results are condensed through dense layers. Finally, the network outputs the probability of the presence of an earthquake.}
	\label{fig:NN3}
\end{figure}
In this problem, we need to extract the fundamental features from time series data to identify seismic events of interest. To achieve this, the first part of our NN consists of a chain of CNNs with a specific progression in the number of neurons. The network starts with a small number of neurons (16) and a minimal kernel size of 1 point, and gradually increases to 128 neurons in the final CNN layer, which uses a window size of 5 points.\\
This particular configuration is designed to capture both fine-grained details, with smaller kernels and fewer neurons in the initial layers, and larger-scale patterns, with wider kernels and more neurons in the deeper layers. This combination provides that the NN can identify both short-term fluctuations, which may correspond to small features of seismic events, and long-term trends, which often characterize the overall event dynamics or teleseismic events that usually have a longer duration filling our entire observation window of 10 min. In Figure \ref{fig:NN3} we can see a representative diagram of the entire structure of the network.\\
At this point, after the last CNN, we introduce an LSTM layer to analyze all extracted features and identify the most relevant groups that reveal the presence of seismic events. To enhance the network's performance, we use a skip connection combined with an attention layer and an addition layer, followed by a final LSTM layer. The skip connection allows the network to bypass certain intermediate layers, enabling the preservation of important features from earlier stages of the network. This helps reducing the risk of vanishing gradients and ensures that relevant information is not lost as it propagates through the deeper layers. Additionally, skip connections can improve learning efficiency by facilitating the flow of gradients during backpropagation.\\
The attention layer is included to focus on the most relevant features extracted by the previous layers \cite{vaswani2023attentionneed}, by assigning different weights to different features, highlighting those that are most indicative of the presence of seismic events.\\
The final LSTM layer processes this enhanced feature representation, capturing temporal dependencies and further refining the information before passing it to the fully connected layers. This combination provides that the network is both robust and effective in detecting seismic events, even in complex and noisy datasets. The output is then passed through a typical sequence of fully connected layers to gradually reduce the number of neurons down to 1. \\
The final Dense layer uses a sigmoid activation function, as this is a classification problem. The sigmoid function maps the output to a range between 0 and 1, making it suitable for predicting the probability of the presence of a seismic event.\\
The performance of the NN was evaluated using a custom metric combining several standard classification metrics: accuracy, precision, recall, and Area Under the Curve (AUC). The AUC is computed from the Receiver Operating Characteristic curve, which plots the true positive rate against the false positive rate for various classification thresholds, providing a comprehensive view of the model's discriminative performance.\\
These metrics assess the model’s ability to correctly identify seismic events and avoid false detections, as discussed in detail in the literature \cite{powers2020evaluationprecisionrecallfmeasure}. Specifically, \textit{accuracy} measures the overall proportion of correct predictions; \textit{precision} quantifies how many predicted earthquakes are true events; \textit{recall} evaluates the network’s ability to detect actual earthquakes; and \textit{AUC} summarizes the trade-off between true and false positives across thresholds.\\
As mentioned before, we selected 50 events labeled as “1” and 50 events labeled as “0” to create a dataset for testing our network. The best result achieved recorded the highest values across all the metrics previously introduced. Figure \ref{CMatrix} (top) presents the confusion matrix for the test and training data. On the x-axis, the predicted number of events for each of the two classes is shown, while on the y-axis, the actual number of events is displayed. The results demonstrate that we achieved an accuracy between 99\% and 100\% on data that the network had never seen before. However, due to the small sample size of only 100 events, it is not possible to appreciate the first decimal place.
Instead of the test data, we can also evaluate the accuracy on the training dataset (which includes both the training and validation data) to observe the first decimal place. In fact, the best network achieves an accuracy of 99.6\%. As shown in Figure \ref{CMatrix} (bottom), it is important to note that not all positive events are correctly classified.\\
Studying more closely the 4 events that were not correctly classified, shown in Figure \ref{FN}, we can infer, a posteriori, the features it prioritized for earthquake recognition. In fact, the network assigns to only one event a probability very close to the threshold value of 50\% used to discriminate between earthquake and non-earthquake events. This event is also the only one showing clear signs of an earthquake in the signal, with oscillatory patterns that can be visually associated with local seismic events. In contrast, all the other signals exhibit no such evidence, or only minimal indications that become visible after applying a low-pass filter (shown in yellow). It should be noted, however, that the signal fed into the network is the unfiltered one.
This behavior suggests that the network has successfully learned the relevant features to discriminate the presence or absence of an earthquake, having misclassified only four examples during training, while achieving perfect classification scores on both the validation and test datasets. In the future, it may be worth considering providing only the probabilities estimated by the network, leaving the final choice of the threshold for determining the presence of an earthquake to subsequent post-processing analyses.\\
This is a preliminary attempt to build a classifier designed to complement a Data Acquisition system for seismology. As previously mentioned, even at this level of accuracy, we expect to identify new potential events that can be studied together with the co-located seismometers.

\section{ Conclusions and Future Prospects}

In conclusion, we demonstrated how the frequency of a RLG can be estimated in 10 ms time using NN, which outperforms the standard ST algorithm. This network achieves up to twice the precision within GINGERINO's typical frequency range and up to four times the precision for higher frequencies near 100\,Hz. The ability to accurately reconstruct frequencies, even for Sagnac biases smaller than those currently accessible, opens the possibility of exploring a wider range of projection angles $\theta$ with respect to Earth’s rotation axis, and also enables the use of gyroscopes with smaller cavity without losing the capability of fast signal reconstruction. This flexibility is particularly valuable for future studies aimed at characterizing the full rotational response of a given site. Moreover, this capability enables the real-time generation of masks that flag disturbances affecting the apparatus, particularly those related to laser dynamics.\\
Based on these results, we developed a second NN for the classification of seismic events. This model achieved an accuracy between 99\% and 100\% on a test set of 100 events not used during training. However, the current dataset lacks an isotropic and homogeneous distribution of local and regional earthquakes. Future efforts will focus on incorporating events from currently underrepresented geographic areas, see Figure~\ref{fig:Mappa}. As with disturbance signals, the network will first be tested on these new events and then retrained, continually improving its generalization and reliability.\\
To ensure robustness, the dataset will be expanded with labeled non-seismic events, such as laser disturbances and glitches, under the class \textit{No earthquake}. In addition, teleseisms will be simulated based on known parameters such as magnitude and source characteristics, thereby avoiding the need to wait for a sufficient number of real examples and allowing us to augment the dataset synthetically while improving the network’s ability to generalize across different types of seismic signals. The network will then be retrained from the best performing weights and evaluated to ensure that its accuracy is preserved.\\

\appendices

\section{Tests on Single Tone Algorithm}
\label{A1}

\begin{figure}[h]
		\includegraphics[scale=0.07]{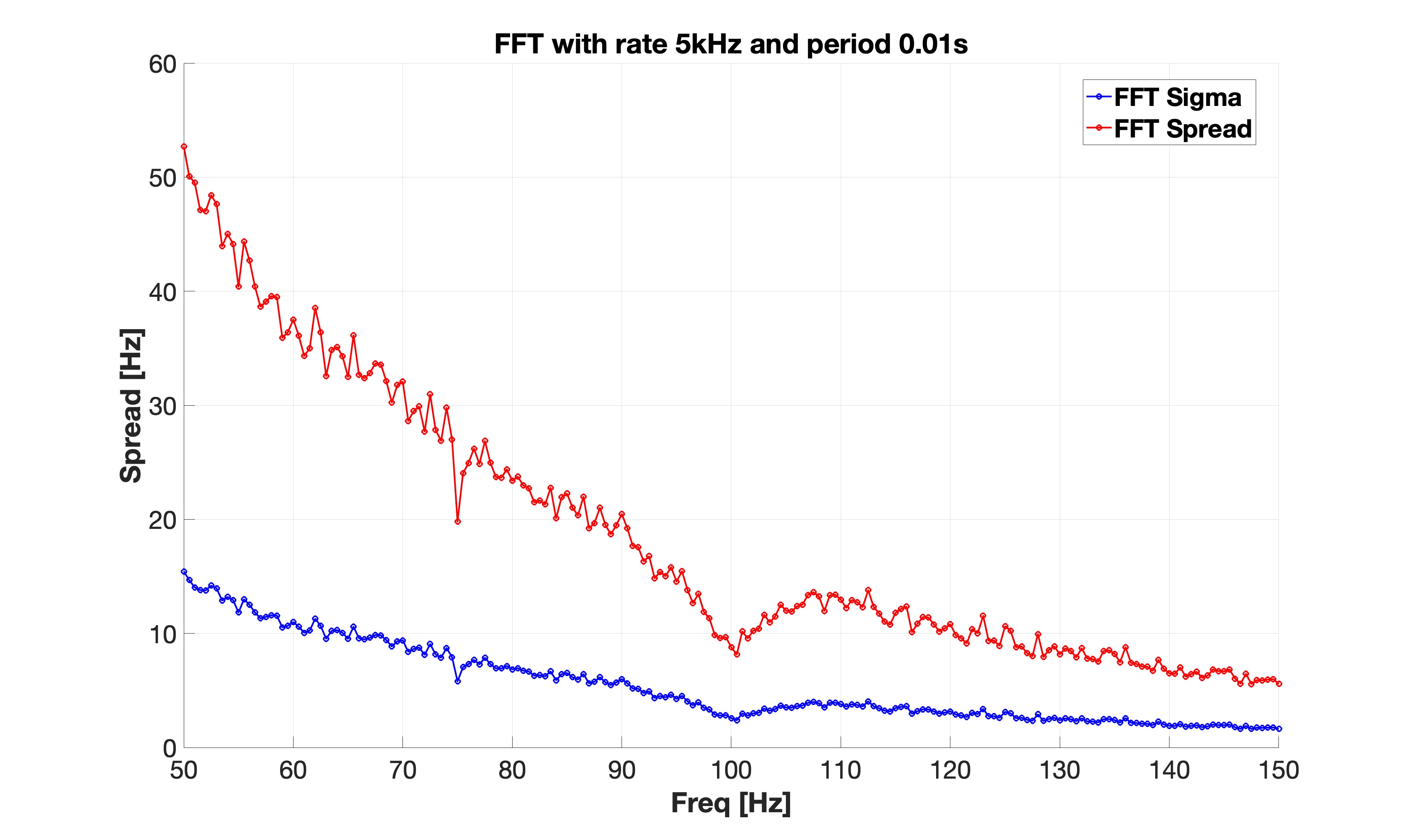}

		\includegraphics[scale=0.07]{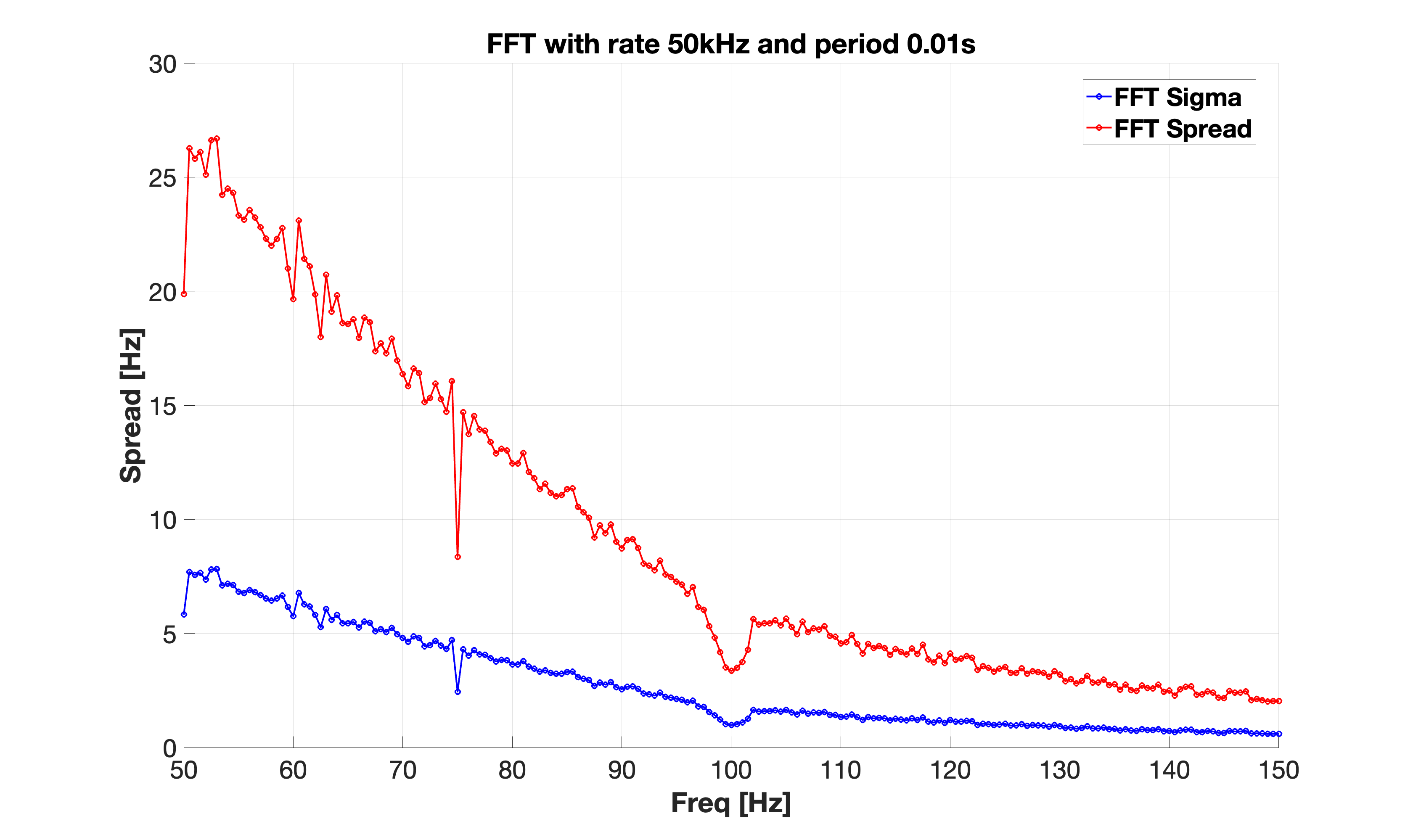}

		\includegraphics[scale=0.07]{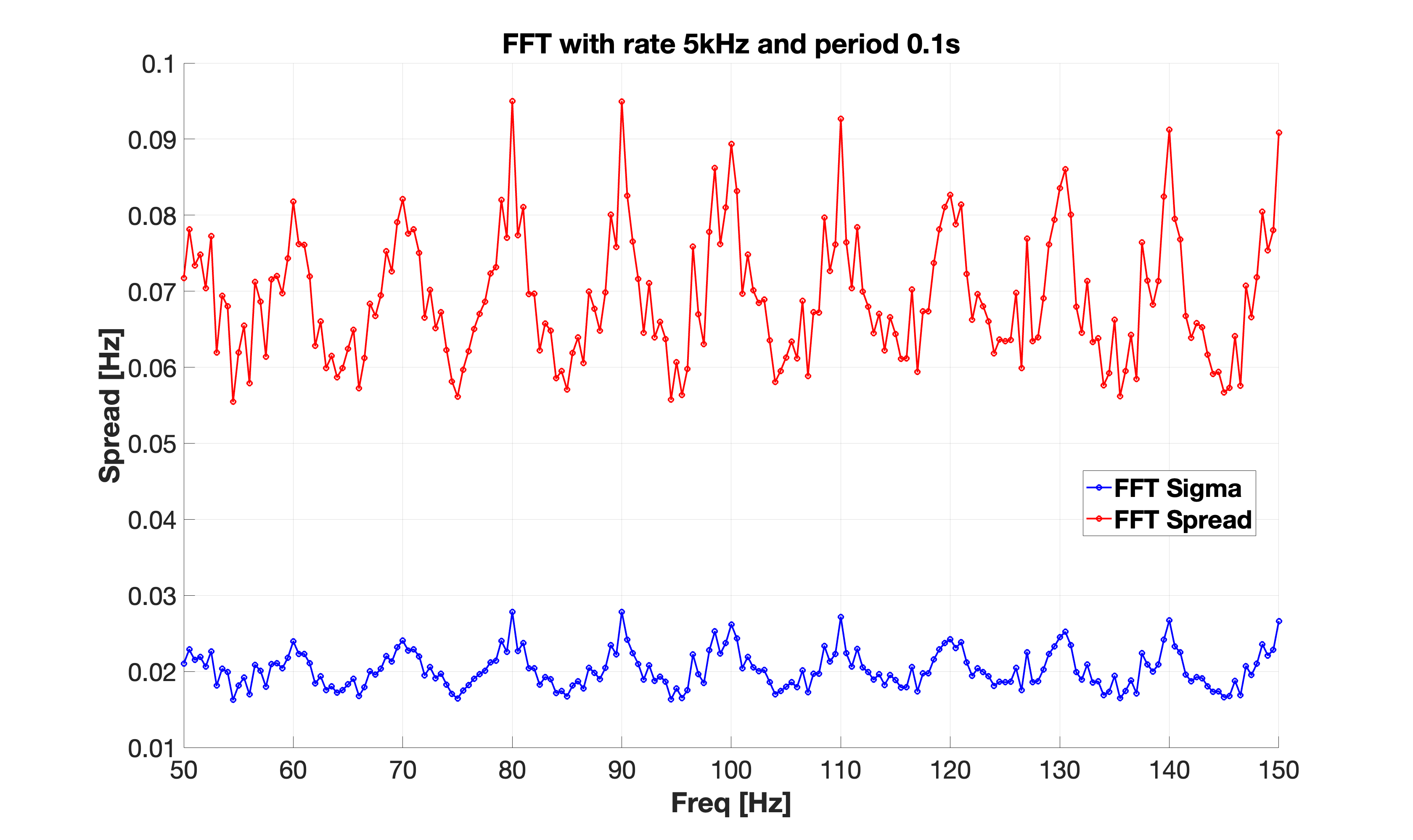}
	\caption{Considering only ST, the spread (red curve) and the standard deviation (blue curve) are shown in the range [50 Hz, 150 Hz]. Top: standard operating conditions (5 kHz, 0.01 s). Middle: increased sampling rate to 50 kHz with the same 0.01 s duration. Bottom: same 5 kHz sampling rate but with a tenfold longer duration.}
	\label{FR5}
\end{figure}
To better understand the behavior of the ST at lower frequencies, we conducted tests exclusively on this method. We generated simulated signals for each target frequency, ranging from 50 Hz to 150 Hz, with increments of 0.5 Hz.
As before, for each of these frequencies, we created 10000 sinusoidal signals and added random noise to simulate measurement uncertainty.\\
Our results showed that at lower frequencies (50–100 Hz), the ST showed higher dispersion, reflected in larger $\sigma$ values and spread, see Figure \ref{FR5}. This can be explained by the fact that the ST has a poorer relative resolution at low frequencies, which means small errors in frequency estimation become more significant. Additionally, spectral leakage is more pronounced at lower frequencies, meaning that the energy of the signal tends to spread over multiple frequency bins, leading to more dispersed frequency estimates. As we increased the frequency of the signals, especially beyond 100 Hz, both $\sigma$ and spread decreased, indicating that the ST produced more accurate and concentrated frequency estimates. The improvement in precision at higher frequencies is demonstrated by the reduced relative error and less significant spectral leakage. Moreover, the impact of the noise is more noticeable at lower frequencies, further contributing to the higher dispersion of the estimates in that range.\\
In summary, the tests revealed that the ST tends to produce more errors when estimating lower frequencies in the presence of noise, due to its poorer relative resolution and greater spectral leakage, while its performance improves significantly at higher frequencies.\\
It is worth noting that in both cases, the results improve significantly, and as the duration increases, the precision improves by an order of magnitude. However, we are limited to a 5 kHz acquisition rate due to the experimental setup and aim to provide data at 0.01-second intervals to deliver frequency as quickly as possible.\\

\printbibliography

\vspace{5mm}

We acknowledge the "Istituto Nazionale di Fisica Nucleare" for funding the experiment, and we express our deepest appreciation to Prof. Nicolò Beverini for his fruitful advice, insightful discussions, and valuable lectures.

\end{document}